\documentclass[journal]{IEEEtran}

\usepackage{amsmath} %
\usepackage{amssymb}  %
\usepackage{tensor}
\usepackage{mathtools}
\usepackage{makecell}
\usepackage{siunitx}[=v2]
\usepackage{booktabs}
\usepackage{amssymb}
\usepackage{multirow}
\usepackage{url}
\usepackage{graphicx}
\usepackage{xcolor}
\usepackage[caption=false, font=footnotesize]{subfig}
\usepackage{balance}
\usepackage{xprintlen}

\title{VILENS: Visual, Inertial, Lidar, and Leg Odometry for All-Terrain Legged
Robots}
\author{\IEEEauthorblockN{David Wisth,~\IEEEmembership{Graduate Student 
Member,~IEEE},
		Marco Camurri,~\IEEEmembership{Member,~IEEE},\\ and
		Maurice Fallon,~\IEEEmembership{Senior Member,~IEEE}}\\
	\IEEEauthorblockA{Oxford Robotics Institute,
		University of Oxford, Oxford, UK}
	\thanks{Manuscript received June 28, 2021; revised January 7, 2022.
		Corresponding author: D. Wisth (email: davidw@robots.ox.ac.uk).}}

\newcommand{\etalcite}[2]{#1~\textit{et~al.}~\cite{#2}}

\newcommand{\Figure}{Fig.~}
\newcommand{\Figures}{Figs.~}
\newcommand{\Equation}{Eq.~}
\newcommand{\Equations}{Eqs.~}
\newcommand{\ie}{{i.e.,~}}
\newcommand{\eg}{{e.g.,~}}

\newcommand{\SO}{\mathrm{SO}}
\newcommand{\SOthree}{\SO(3)}
\newcommand{\Real}{\mathbb{R}}
\newcommand{\Realthree}{\Real^{3}}
\newcommand{\expmap}{\mathrm{Exp}}
\newcommand{\logmap}{\mathrm{Log}}

\newcommand{\calA}{{\cal A}}
\newcommand{\calB}{{\cal B}}
\newcommand{\calC}{{\cal C}}

\newcommand{\calI}{{\cal I}}

\newcommand{\calK}{{\cal K}}
\newcommand{\calL}{{\cal L}}

\newcommand{\Force}{\mathbf{f}}
\newcommand{\Torque}{\boldsymbol{\tau}}

\newcommand{\R}{\mathbf{R}}
\newcommand{\Identity}{\mathbf{I}}
\newcommand{\T}{\mathbf{T}}

\newcommand{\residual}{\mathbf{r}}
\newcommand{\transpose}{\mathsf{T}}

\newcommand{\rotvel}{\boldsymbol\omega}
\newcommand{\rotvec}{\boldsymbol\phi}
\newcommand{\rotvectwist}{\boldsymbol\theta}
\newcommand{\tran}{\mathbf{p}}
\newcommand{\vel}{\mathbf{v}}
\newcommand{\bias}{\mathbf{b}}
\newcommand{\gravity}{\mathbf{g}}
\newcommand{\noise}{\boldsymbol\eta}

\newcommand{\States}{\mathcal{X}}
\newcommand{\State}{\boldsymbol{x}}
\newcommand{\Measurements}{\mathcal{Z}}

\newcommand{\Base}{\mathtt{{B}}}
\newcommand{\Camera}{\mathtt{C}}
\newcommand{\Lidar}{\mathtt{L}}
\newcommand{\World}{\mathtt{W}}
\newcommand{\Imu}{\mathtt{I}}
\newcommand{\Contact}{\mathtt{K}}
\newcommand{\Anchor}{\mathtt{{A}}} %

\newcommand{\base}{\mathtt{{B}}}

\newcommand{\world}{\mathtt{{W}}}
\newcommand{\imu}{\mathtt{{I}}}

\newcommand{\Plane}{\boldsymbol{p}}

\newcommand{\Line}{\boldsymbol{l}}

\newcommand{\Landmark}{\boldsymbol{m}} %

\newcommand{\w}{\omega}
\newcommand{\wb}{\boldsymbol{\w}}
\newcommand{\ba} {\bias^a} %
\newcommand{\bg} {\bias^g} %
\newcommand{\bv} {\bias^v} %
\newcommand{\bw} {\bias^\omega} %

\newcommand{\bvi}{\bias^v_i}

\newcommand{\etav}{\noise^v}
\newcommand{\etaw}{\noise^\omega}
\newcommand{\etag}{\noise^g}
\newcommand{\etajp}{\noise^{\alpha}}
\newcommand{\etajv}{\noise^{\dot{\alpha}}}

\newcommand{\tv}{\tilde{\vel}}
\newcommand{\tw}{\tilde{\wb}}

\newcommand{\tvk}{\tilde{\vel}_k}
\newcommand{\twk}{\tilde{\rotvel}_k}

\newcommand{\dep}{\delta \pretwist}
\newcommand{\depij}{\dep_{ij}}

\newcommand{\dkij}{\Delta \pretwist_{ij}}

\newcommand{\dRik}{\Delta \R_{ik}}
\newcommand{\dPhi}{\delta \rotvec}
\newcommand{\dVtheta}{\delta \rotvectwist}
\newcommand{\dt}{\Delta t}
\newcommand{\dR}{\Delta \R}
\newcommand{\dtp}{\Delta\tilde{\pretwist}}
\newcommand{\dtTheta}{\Delta\tilde{\boldsymbol{\Theta}}}

\newcommand{\dtRij}{\Delta\tilde{\R}_{ij}}

\newcommand{\dtpij}{\dtp_{ij}}
\newcommand{\dtvij}{\Delta\tilde{\vel}_{ij}}

\newcommand{\dPhiij}{\delta\rotvec_{ij}}

\newcommand{\jointpos}{{\boldsymbol{\alpha}}}
\newcommand{\jointvel}{{\dot{\jointpos}}}

\newcommand{\fk}[1]{\mathbf{f}_{#1}}
\newcommand{\fkp}{\fk{p}}

\newcommand{\kJac}[1]{\mathbf{J}_{#1}}
\newcommand{\Jacp}{\kJac{p}}
\newcommand{\JacR}{\kJac{R}}

\newcommand{\sumjmo} [1]{\sum_{k=i}^{j-1}\left[ #1 \right]}

\newcommand{\Exp}[1]{\expmap \left( #1 \right)}
\newcommand{\Log}[1]{\logmap \left( #1 \right)}

\newcommand{\defeq}{\triangleq}

\DeclareMathOperator*{\argmax}{arg\,max}
\DeclareMathOperator*{\argmin}{arg\,min}

\newcommand{\jp}{\jointpos}
\newcommand{\jv}{\jointvel}
\newcommand{\Jacpp}[1]{\mathbf{J}_p(#1)}
\newcommand{\tjp}{\tilde{\jp}}
\newcommand{\tjv}{\tilde{\jv}}
\newcommand{\partderivbig}[1]{\frac{{\partial}}{{\partial}#1}}
\newcommand{\hessian}[1]{\mathbf{H}_p(#1)}
\newcommand{\gaussian}[2]{\mathcal{N}(#1,\,#2)}

\newcommand{\Zero}{\mathbf{0}}
\newcommand{\eye}{{\mathbf I}}

\newcommand{\acc}{\mathbf{a}}
\newcommand{\tacc}{\tilde{\acc}}

\newcommand{\tranpert}{\delta\tran}
\newcommand{\velpert}{\delta\vel}
\newcommand{\pretwist}{\boldsymbol{\kappa}}
\newcommand{\rhopert}{\delta\pretwist}

\newcommand{\Cov}{\mathbf{\Sigma}}
\newcommand{\InfoMat}{\mathbf{\Omega}}

\newcommand{\measimu}{\mathcal{I}}

\newcommand{\preintRmeas}{\Delta\tilde\R}
\newcommand{\preintVmeas}{\Delta\tilde\vel}
\newcommand{\preintPmeas}{\Delta\tilde\tran}

\newcommand{\Amat}{\mathbf{A}}
\newcommand{\Bmat}{\mathbf{B}}
\newcommand{\Cmat}{\mathbf{C}}

\makeatletter
\def\underbracex#1#2{\mathop{\vtop{\m@th\ialign{##\crcr
   $\hfil\displaystyle{#2}\hfil$\crcr
   \noalign{\kern3\p@\nointerlineskip}%
   #1\crcr\noalign{\kern3\p@}}}}\limits}

\def\underbracea{\underbracex\upbracefilla}

\def\upbracefilla{$\m@th \setbox\z@\hbox{$\braceld$}%
  \bracelu\leaders\vrule \@height\ht\z@ \@depth\z@\hfill
\kern\p@\vrule \@width\p@\kern\p@\vrule \@width\p@\kern\p@\vrule \@width\p@
$}

\def\upbracefillb{$\m@th \setbox\z@\hbox{$\braceld$}%
\vrule \@width\p@\kern\p@\vrule \@width\p@\kern\p@\vrule \@width\p@\kern\p@
 \leaders\vrule \@height\ht\z@ \@depth\z@\hfill\bracerd
  \braceld\leaders\vrule \@height\ht\z@ \@depth\z@\hfill
\kern\p@\vrule \@width\p@\kern\p@\vrule \@width\p@\kern\p@\vrule \@width\p@
$}

\def\underbraced{\underbracex\upbracefilld}
\def\upbracefilld{$\m@th \setbox\z@\hbox{$\braceld$}%
\vrule \@width\p@\kern\p@\vrule \@width\p@\kern\p@\vrule \@width\p@\kern\p@
 \leaders\vrule \@height\ht\z@ \@depth\z@\hfill\bracerd
  \braceld\leaders\vrule \@height\ht\z@ \@depth\z@\hfill
 \braceru$}

\makeatother

\makeatletter
\let\NAT@parse\undefined
\makeatother
\usepackage{hyperref}
\usepackage{tikz}
\usetikzlibrary{shapes,arrows, shapes.multipart}

\newcommand{\Seemuhle}{Seem\"uhle }

\begin{document}

\onecolumn
\thispagestyle{empty}

\hspace{3cm}
\begin{center}
This paper has been accepted for publication in \emph{IEEE Transactions 
on Robotics} (T-RO).\\

\hspace{1cm}

DOI: 
\href{https://doi.org/10.1109/TRO.2022.3193788}{10.1109/TRO.2022.3193788}\\

IEEE Explore: \url{https://ieeexplore.ieee.org/document/9852710} \\

\hspace{1cm}

Please cite the paper as: \\

\hspace{1cm}

D. Wisth, M. Camurri and M. Fallon,\\
``VILENS: Visual, Inertial, Lidar, and Leg Odometry for All-Terrain Legged 
Robots,'' \\
in IEEE Transactions on Robotics, vol. 39, no. 1, pp. 309-326, Feb. 
2023, \\
doi: 10.1109/TRO.2022.3193788.

\end{center}
\twocolumn

\bstctlcite{library:BSTcontrol}

\markboth{IEEE TRANSACTIONS ON ROBOTICS, VOL. 39, NO. 1, FEBRUARY 2023}%
{IEEE TRANSACTIONS ON ROBOTICS, VOL. 39, NO. 1, FEBRUARY 2023}
\maketitle

\begin{abstract}
We present visual inertial lidar legged navigation system (VILENS), an odometry
system for legged robots based on factor graphs. The key novelty is the tight
fusion of four different sensor modalities to achieve reliable operation when
the individual sensors would otherwise produce degenerate estimation. To
minimize leg odometry drift, we extend the robot’s state with a linear velocity
bias term, which is estimated online. This bias is observable because of the
tight fusion of this preintegrated velocity factor with vision, lidar, and 
inertial measurement unit (IMU)
factors. Extensive experimental validation on different ANYmal quadruped robots
is presented, for a total duration of \SI{2}{\hour} and \SI{1.8}{\kilo\meter}
traveled. The experiments involved dynamic locomotion over loose rocks, slopes,
and mud, which caused challenges such as slippage and terrain deformation. 
Perceptual challenges included
dark and dusty underground caverns, and open and
feature-deprived areas. We show an average improvement of \SI{62}{\percent}
translational and \SI{51}{\percent} rotational errors compared to a
state-of-the-art loosely coupled approach. To demonstrate its robustness, VILENS
was also integrated with a perceptive controller and a local path planner.
\end{abstract}


\IEEEpeerreviewmaketitle

\section{Introduction}
\IEEEPARstart{T}{he} increased maturity of quadruped robotics has been
demonstrated in initial industrial deployments, as well as impressive results
achieved by academic research. State estimation plays a key role in field
deployment of legged machines: Without an accurate estimate of its location and
velocity, the robot cannot build up a useful representation of its environment,
or plan and execute trajectories to reach desired goal positions.

Most legged robots are equipped with a high-frequency (\SI{>250}{\hertz})
proprioceptive state estimator for control and local mapping purposes. These are
typically implemented as nonlinear filters fusing high-frequency signals, such 
as
kinematics and inertial measurement unit (IMU) \cite{Bloesch2013} 
\cite{Bloesch2017tsif} \cite{hartleyIJRR}.
In ideal conditions (\ie high friction, rigid terrain, slow speeds), these
estimators have a limited (yet unavoidable) drift that is acceptable for local
mapping and control.

However, deformable terrain, leg flexibility, and foot slippage can degrade
estimation performance up to the point where local terrain reconstruction is
unusable and multistep trajectories cannot be executed, even over short ranges.
This problem is more evident when a robot is moving dynamically and
can be the limiting factor when crossing rough terrain. This estimate is also
affected by modeling errors, such as inaccurate leg lengths or nonzero
contact point size.

Recent works have attempted to improve kinematic-inertial estimation accuracy by
reducing the convergence time using invariant observer design \cite{hartleyIJRR}
or by detecting unstable contacts and reducing their influence on the overall
estimation \cite{Camurri2017,Jenelten2019,jemin2021}.

Other approaches have
incorporated
additional exteroceptive sensing into the estimator to help reduce the pose
error. These included either tightly coupled methods fusing camera, IMU, and
kinematics \cite{Hartley2018b} or loosely coupled methods combining lidar in
addition to the other sensors \cite{Camurri2020, Khattak2020}. These approaches
model the contact locations as being fixed and affected only by
Gaussian noise. Both assumptions are broken when there is nonrigid terrain,
kinematic chain flexibility, or mild but repeated foot slippage. When these
occur, fusion with exteroceptive sensors becomes nontrivial.

In our work, we aim to fuse all four sensor modalities (IMU, kinematics,
lidar, camera) in a tightly coupled
fashion, with particular focus on the proper integration of leg kinematics in
presence of non-ideal contacts, when slippage or terrain deformation occurs.

\begin{figure}
 \centering
 \includegraphics[width=0.49\columnwidth]{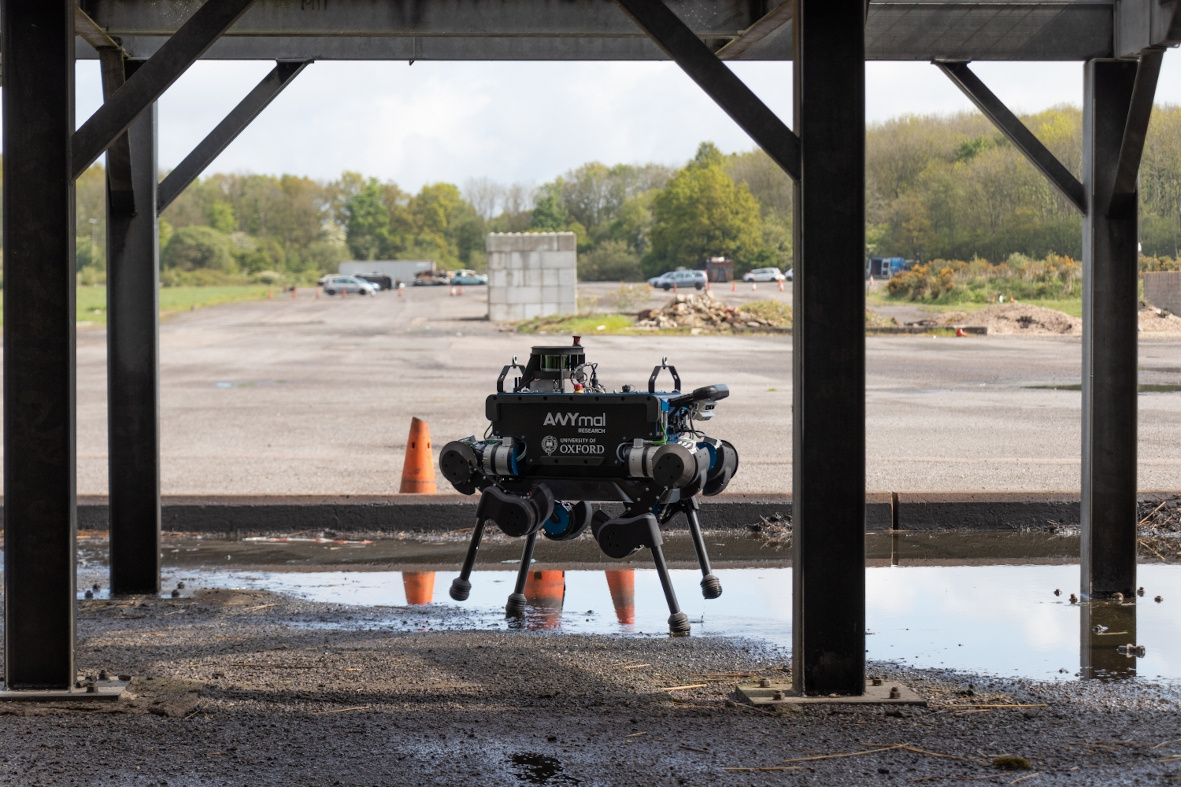}
 \includegraphics[width=0.49\columnwidth]{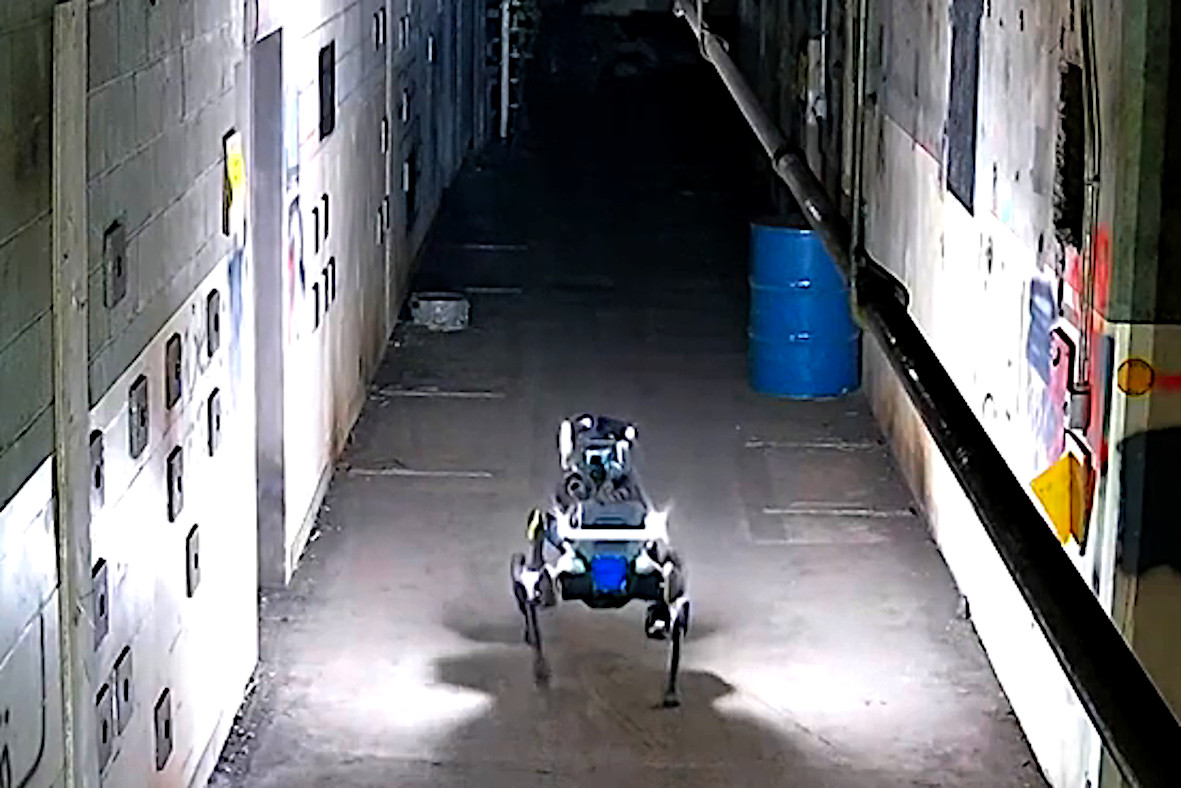}
 \\
 \vspace{0.011428571\columnwidth}
 \includegraphics[width=0.49\columnwidth]{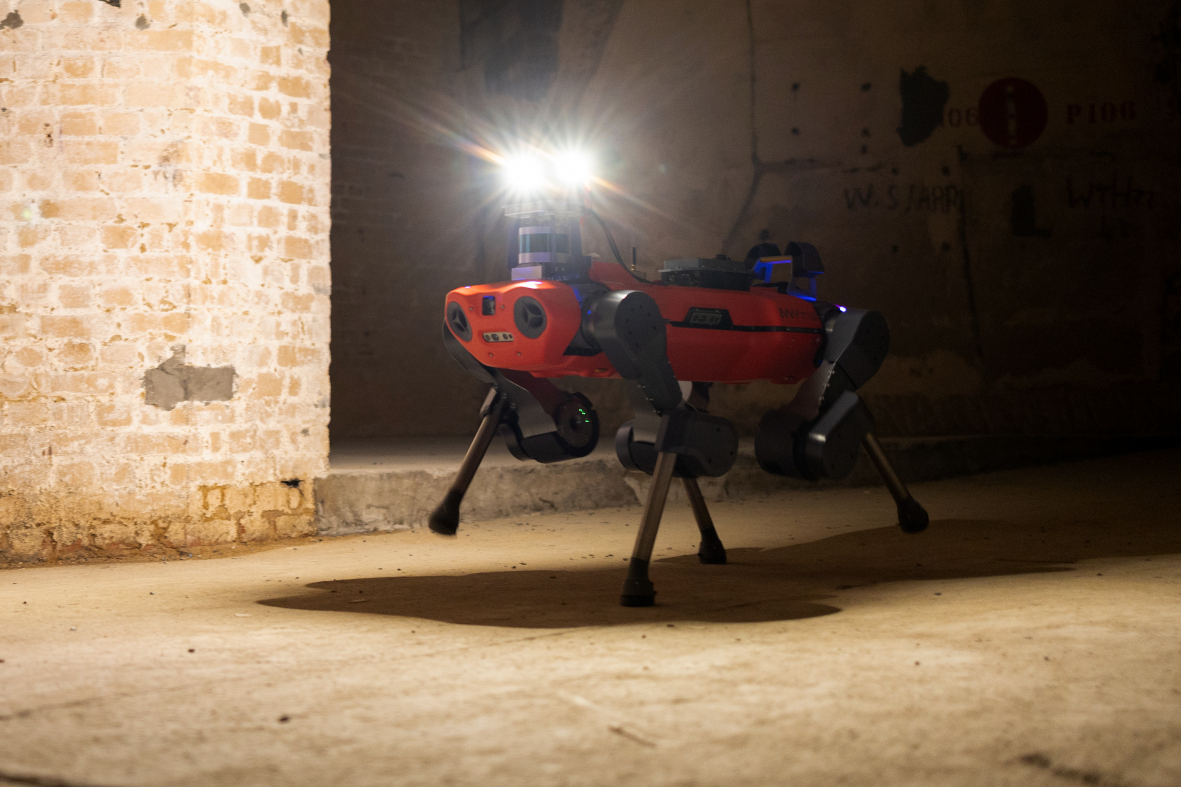}
 \includegraphics[width=0.49\columnwidth]{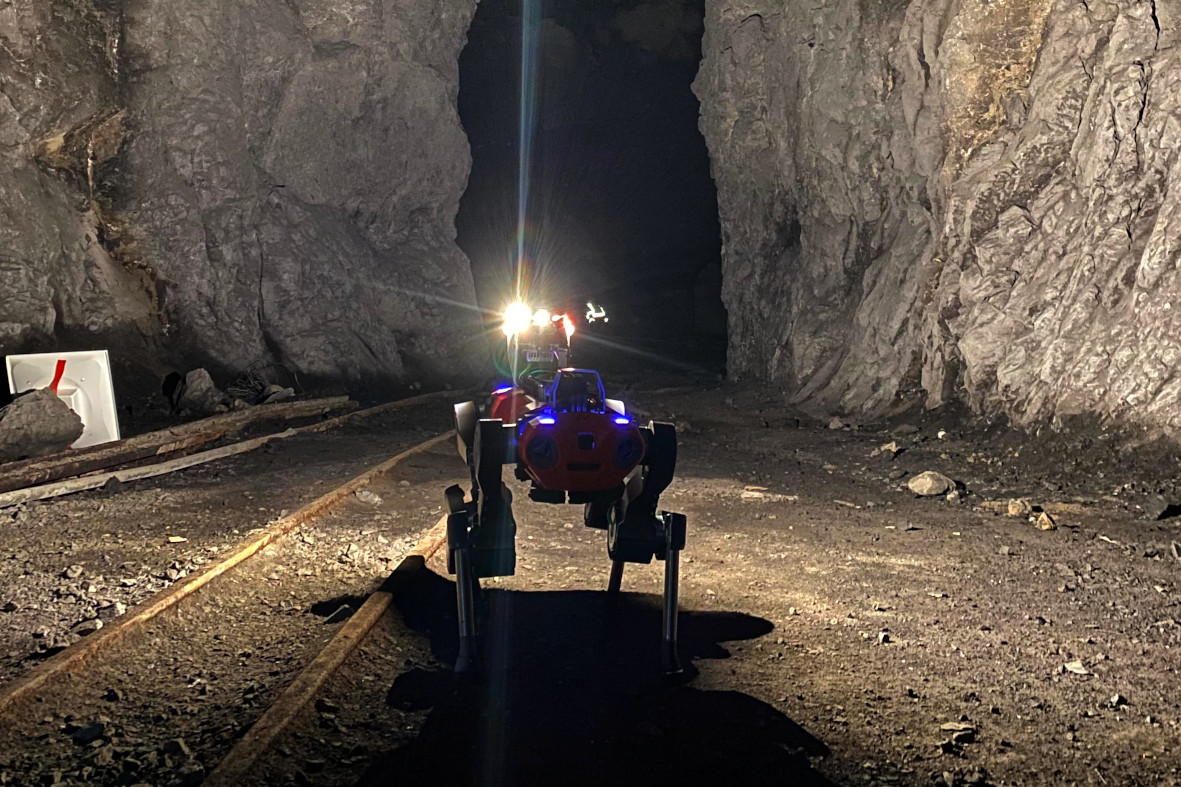}
 \caption{VILENS has been tested on a variety of platforms.  \textit{Top-Left:}
ANYmal B300 at the Fire Service
College in Moreton-On-Marsh (UK).
\textit{Top-Right:} ANYmal B300 modified for the DARPA SubT
Challenge (Urban Circuit) in Olympia (Washington, USA)
\cite{Tranzatto2021cerberus}.
\textit{Bottom-Left:} ANYmal C100 in a limescale mine in Wiltshire (UK).
\textit{Bottom-Right:}ANYmal C100 in an abandoned mine in \Seemuhle
(Switzerland).}
\label{fig:platforms-overview}
\end{figure}

\subsection{Motivation}
\label{subsection:Motivation}
Our work is motivated by the challenges and limitations of state estimation for
the deployment of legged robots in extreme environments, as illustrated in
\Figure \ref{fig:platforms-overview}. The DARPA Subterranean (SubT) Challenge
\cite{Tranzatto2021cerberus}
involves the deployment of a team of autonomous robots navigating unknown, dark,
and unstructured underground environments. In such scenarios, individual sensor
modalities can fail (\eg due to camera blackouts or degenerate geometries for
lidar), so robust sensor fusion is paramount. Additional requirements include
limited computational budget and the need for high-frequency output to update
the local footstep planner. For these reasons, we aim to use all of the sensors
available on the robot (IMU, kinematics, lidar, and camera) to form constraints
for lightweight sliding window optimization, as it can be more accurate than
filter-based approaches \cite{Dellaert2017}.

\begin{figure}
\centering
\includegraphics{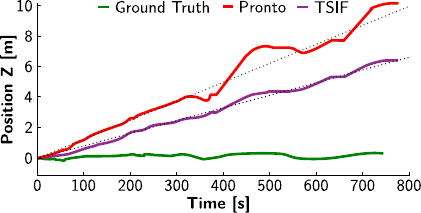}
\caption{Comparison between estimated robot elevation by Pronto
\cite{Camurri2017} (red) and TSIF \cite{Bloesch2017tsif} (purple)
kinematic-inertial state estimators, against ground truth (green) on the SMR
experiment (see Section \ref{sec:datasets}). Despite local fluctuations, the
drift has a characteristic linear growth for a particular gait and terrain
type. For example, between \SI{350}{\second} and \SI{450}{\second} the robot
walks over soft gravel, increasing the drift rate.}
\label{fig:position-drift-motivation}
\end{figure}

When both lidar and camera fail at the same time, IMU integration alone would
rapidly lead to divergence. Leg kinematics can prevent this, but special care
has to be taken to fuse it with the other sensors in an effective way,
especially in the presence of foot impacts and terrain deformation. \Figure
\ref{fig:position-drift-motivation} shows this effect on the ANYmal B300 robot
traveling over various terrains. Two different kinematic-inertial estimators
suffer from continuous elevation drift, which is locally approximately linear
(see dashed lines in \Figure \ref{fig:position-drift-motivation}).
\etalcite{Fahmi}{Fahmi2021} demonstrated that this drift can be caused by
nonrigid, and nonstatic interaction of the legs and terrain during contact
events.

\begin{figure}
\centering
\hspace{-0.02\columnwidth}%
\subfloat[]{
\includegraphics[width=0.24\columnwidth]{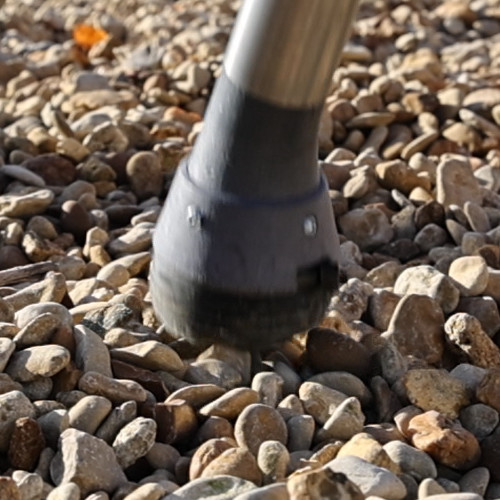}%
}%
\subfloat[]{
\includegraphics[width=0.24\columnwidth]{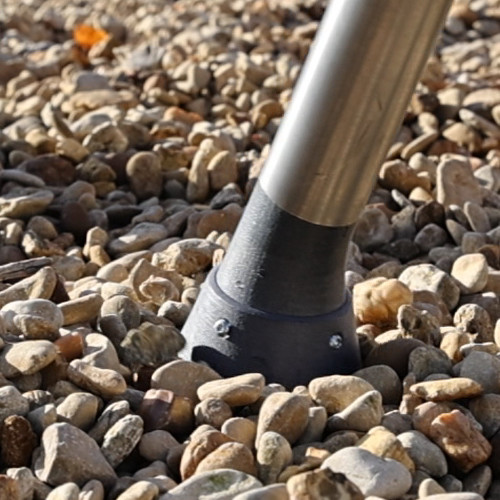}%
}%
\subfloat[]{
\includegraphics[width=0.24\columnwidth]{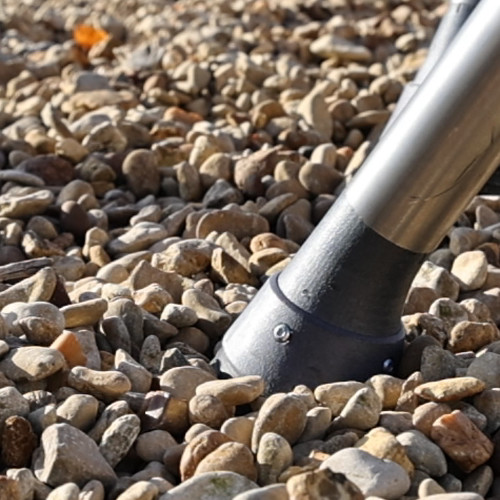}%
}%
\subfloat[]{
\includegraphics[width=0.24\columnwidth]{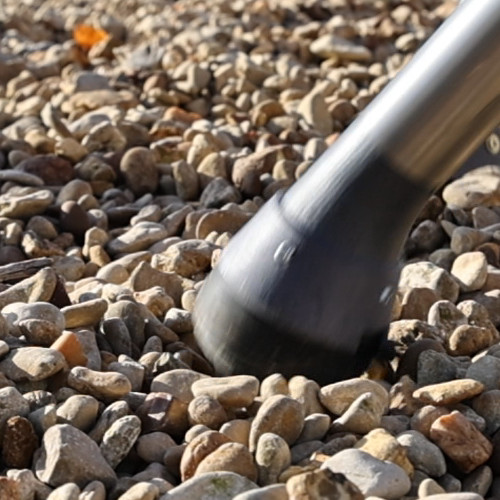}%
}%
\caption{An example of a foot contact sequence while trotting on gravel. After
the foot touches the ground (a), both the terrain and robot's rubber foot
deform as the controller increases the applied force (b). During the stance
phase, the contact point changes as the foot rolls over its hemispherical
profile (c) before finally breaking contact (d).}
\label{fig:foot-contact-sequence}
\end{figure}

An example of a foot contact event on soft gravel is shown in \Figure
\ref{fig:foot-contact-sequence}. The contact point was nonstatic throughout
the entire sequence, violating the key assumption of most leg odometry
algorithms. This is mainly due to deformation of the ground and rubber foot, as
well as the nonzero contact point size (the foot has a hemispherical profile).
This can be considered as a systematic modeling error. Drift will be
accumulated each time the robot steps on this terrain, leading to a biased
estimate.

One approach would be to further model the dynamic properties of the robot, such
as torque-dependent bending \cite{Koolen2016}, or to model the terrain directly
within the estimator. However, this would be robot specific and terrain
dependent -- improving performance in one situation but degrading it elsewhere.
Additionally, threshold-based methods only reject the most significant slippage
or deformation events and ignore the small error accumulated with each footstep.

Inspired by the IMU bias estimation and preintegration methods from
\cite{Forster2017}, we instead propose to extend the state with a velocity bias
term to estimate and reject these effects. This bias is observable when doing
tight fusion with exteroceptive sensors (see Appendix
\ref{sec:observability-analysis}). This novel leg odometry factor computes a
velocity measurement from kinematic sensing, preintegrates it, and estimates its
bias to compensate for the characteristic drift of the leg odometry on slippery
or deformable ground.

\subsection{Contributions}
\noindent This paper makes the following contributions, significantly extending
our previous work \cite{Wisth2020icra}:
\begin{enumerate}
 \item A novel factor that incorporates joint kinematic measurements. In
contrast to the work in \cite{Wisth2020icra}, this factor computes base 
velocities from the
joint states, rather than taking them from an external filter.
This avoids potential correlation due to double IMU usage and allows better
modeling of the noise associated with the kinematics. A full derivation of the
noise propagation is also provided in the Appendix.
\item Support and testing on a wider range of sensor modalities, including
fisheye cameras and lidar. To the best of our knowledge, this is the first
smoothing algorithm to combine inertial, legged, visual, and lidar data into a
single factor graph. In particular, in addition to the high-frequency lidar
feature tracking from \cite{Wisth2020lidar}, we also introduce a lidar odometry
factor for low frequency accurate lidar registration. The wide support of
complementary sensor modalities is the key to operation in adverse operational
conditions, such as the DARPA SubT Challenge.
 \item Extensive evaluation in a broad set of scenarios including different
models of the ANYmal quadruped. In contrast to the work in 
\cite{Wisth2020icra}, which was
validated only on datasets, we tested the algorithm onboard the robot and
integrated it with both a dynamic perceptive controller \cite{Gangapurwala2020}
and an autonomous exploration planner \cite{Dang2020gbplanner}.
\end{enumerate}

The remainder of this article is presented as follows: in Section
\ref{sec:related-work} we review the literature on legged state estimation with
a focus on sensor fusion via smoothing methods and contact modeling; Section
\ref{sec:problem-statement} formally defines the problem addressed by the paper
and provides the required mathematical background; Section
\ref{sec:twist-factors} describes the factors used in our proposed graph
formulation; Section \ref{sec:implementation} presents the implementation
details of our system; Section \ref{sec:results} presents the experimental
results and discussion; Section \ref{sec:conclusions} concludes with final
remarks.

\section{Related Work}
\label{sec:related-work}
We define multisensor state estimation as the joint processing of multiple
proprioceptive and/or exteroceptive sensors to estimate the pose and velocity of
a mobile robot \cite{barfoot2017book}. Multi-sensor estimation can be divided
into two broad categories: \emph{filtering} and \emph{smoothing}.
Filtering methods such as MSC-KF \cite{msckf} restrict
the inference process to the latest state of the system, while
smoothing methods also estimate all or part of the past states. In Section
\ref{sec:smoothing-legged} we describe smoothing methods applied to multisensor
estimation on legged robots. In Section
\ref{sec:handling-contact-deformation-and-slippage} we review the relevant
methods for modeling nonlinearities at the contact point during leg odometry.

\subsection{Multi-Sensor Smoothing on Legged Robots}
\label{sec:smoothing-legged}
Multi-sensor smoothing for legged robots typically involves the fusion of IMU,
leg odometry, and visual tracking within a probabilistic graphical model
framework, such as factor graphs. \etalcite{Hartley}{Hartley2018a} proposed the
first method to incorporate leg odometry into a factor graph. They extended the
floating base state with the feet contact locations and defined two new factors
to incorporate forward kinematics, where zero velocity constraints on the
contact point of each foot were imposed. The two new factors were then combined
with the preintegrated IMU factor from \cite{Forster2017} and relative pose
measurements from the SVO visual odometry system \cite{Forster2017svo}.  In
\cite{Hartley2018b}, the same authors extended this work to support multiple
footsteps in the same kinematics factor. Both works were demonstrated on the
Cassie bipedal robot through experiments in controlled environments lasting
\SI{<2}{\minute}.

\etalcite{Fourmy}{Fourmy2020} proposed a factor-graph based state estimation
system for the Solo12 quadruped robot \cite{Grimminger2020ral}. It fused both
kinematic and dynamic information to estimate the base frame of the robot and
performed online calibration of the offset between the base frame and center of
mass. The authors claimed this was important for control since the center of
mass of their robot was not known precisely from the CAD model. However, only
very limited experimental results were presented, where the robot's torso moved
while the feet were stationary. In the related field of motion planning,
\etalcite{Xie}{Xie2020} modeled both the kinematics and dynamics of a 3 DoF
manipulator using factor graphs. This approach took into account dynamics,
contact forces, and joint actuation limits.

In our prior work \cite{Wisth2019}, we proposed a tightly coupled
visual-inertial-legged system based on the iSAM2 solver \cite{Kaess2012} running
on the ANYmal robot. The method tracked visual features from a RealSense D435i
stereo camera and optimized them as landmarks in a factor graph. Leg
odometry was only loosely coupled with relative pose factors formed using the
internal state estimator running on the robot \cite{Bloesch2017tsif}. This
method was demonstrated through outdoor experiments in urban and industrial
scenarios.

All these works were based on the assumption of a stationary point of foot
contact. This assumption is violated every time there are slippages, or
deformations of the leg and/or the ground. Contact detection methods can help to
reject sporadic slippage or deformation events. However, when these occur
regularly, they need to be modeled.

\subsection{Modeling Contact Deformation and Slippage}
\label{sec:handling-contact-deformation-and-slippage}

In legged robotics, slippage and/or deformation have typically been addressed by
assuming the contact location of a stance foot is entirely static throughout the
stance period (yet affected by Gaussian noise). Thus, the main focus has been on
detecting and ignoring the feet that are not in fixed contact with the ground.
This is a relatively simple task when a foot is equipped with force/torque
sensors. In this case, a high vertical component of the measured force would
imply that the contact force is within the friction cone and, therefore,
nonslipping. However, residual errors due to model uncertainty or deformation
might persist. \etalcite{Fahmi}{Fahmi2021} have shown that incorrect contact
detections on soft ground (\ie detecting a ``rigid'' contact while the leg and
ground are still deforming) are a key contributor to leg odometry drift.

\etalcite{Bloesch}{Bloesch2013} proposed an Unscented Kalman Filter design that
fused IMU and differential kinematics. The approach used a threshold on the
Mahalanobis distance of the filter innovation to infer velocity measurement
outliers (caused by misclassified contact legs) which were then ignored. More
recently, the idea was extended by \etalcite{Kim}{jemin2021} where the
threshold on the Mahalanobis distance was replaced with a tunable threshold on
the contact foot velocity estimated from the previous state.

For systems without feet sensors, more sophisticated methods were proposed.
\etalcite{Hwangbo}{Hwangbo2016} presented a probabilistic approach where
information from kinematics, differential kinematics, and dynamics were fused
within a Hidden Markov Model (HMM). This approach was later integrated
with a dynamic trotting controller and demonstrated on the ANYmal robot walking
on ice \cite{Jenelten2019}.

Instead of an HMM, \etalcite{Bledt}{bledt2018icra} proposed to fuse information
from kinematics and dynamics, as well as additional input from the controller's
gait cycle within a Kalman filter. Their work was demonstrated on the Cheetah 3
robot walking on rubble. Both \cite{Hwangbo2016} and \cite{bledt2018icra}
focused more on detecting the contact as early as possible for control purposes,
rather than determining the contact periods that would minimize the state
estimation error.

\etalcite{Camurri}{Camurri2017} proposed a contact detector that learned the
optimal force threshold to detect a foot in contact for a specific gait, and an
impact detector that adapted the measurement covariance online to reject
unreliable measurements.

In our previous work \cite{Wisth2020icra}, we used this
approach to fuse each leg's contribution into a single velocity measurement for
our proposed factor graph method. The contact nonlinearities were modeled as a
bias term on the linear velocity measurements from leg odometry. This could
reduce the inconsistency between leg and visual odometry and provide a more
robust pose and velocity estimate. However, the system was still dependent on an
external filter to access the velocity estimates from the kinematics.
Furthermore, lidar, which is now a common sensor on quadruped robots, was not
used for estimation.

In contrast to the work in \cite{Wisth2020icra}, in this work, we compute the 
velocity
measurements from kinematics internally, instead of receiving them from an
external filter. This has several benefits: It eliminates the double usage of
the IMU signal (from the filter and from the preintegrated IMU factor) which
breaks the independent measurement assumption; it properly models the error
propagation from the joints to the feet, thereby estimating the covariance from
leg kinematics more accurately; it allows for better integration, as the
optimized IMU and kinematics biases are directly accessible from the
kinematics module, whereas the external filter did not. Finally, it simplifies
the overall estimation architecture, eliminating the dependency on an external
component. We also introduce a lidar registration factor with local submapping
and tested our algorithm online and onboard the robot in conjunction with a
perceptive controller and a local path planner (as opposed to operation in post
processing).

\section{Problem Statement}
\label{sec:problem-statement}
Our objective is to estimate the history of poses and velocities of a legged
robot equipped with a combination of sensors, including: cameras (mono or
stereo), IMUs, lidars, and joint sensors (encoders and torque sensors).

\begin{figure}
\centering
\includegraphics[height=5.0cm]{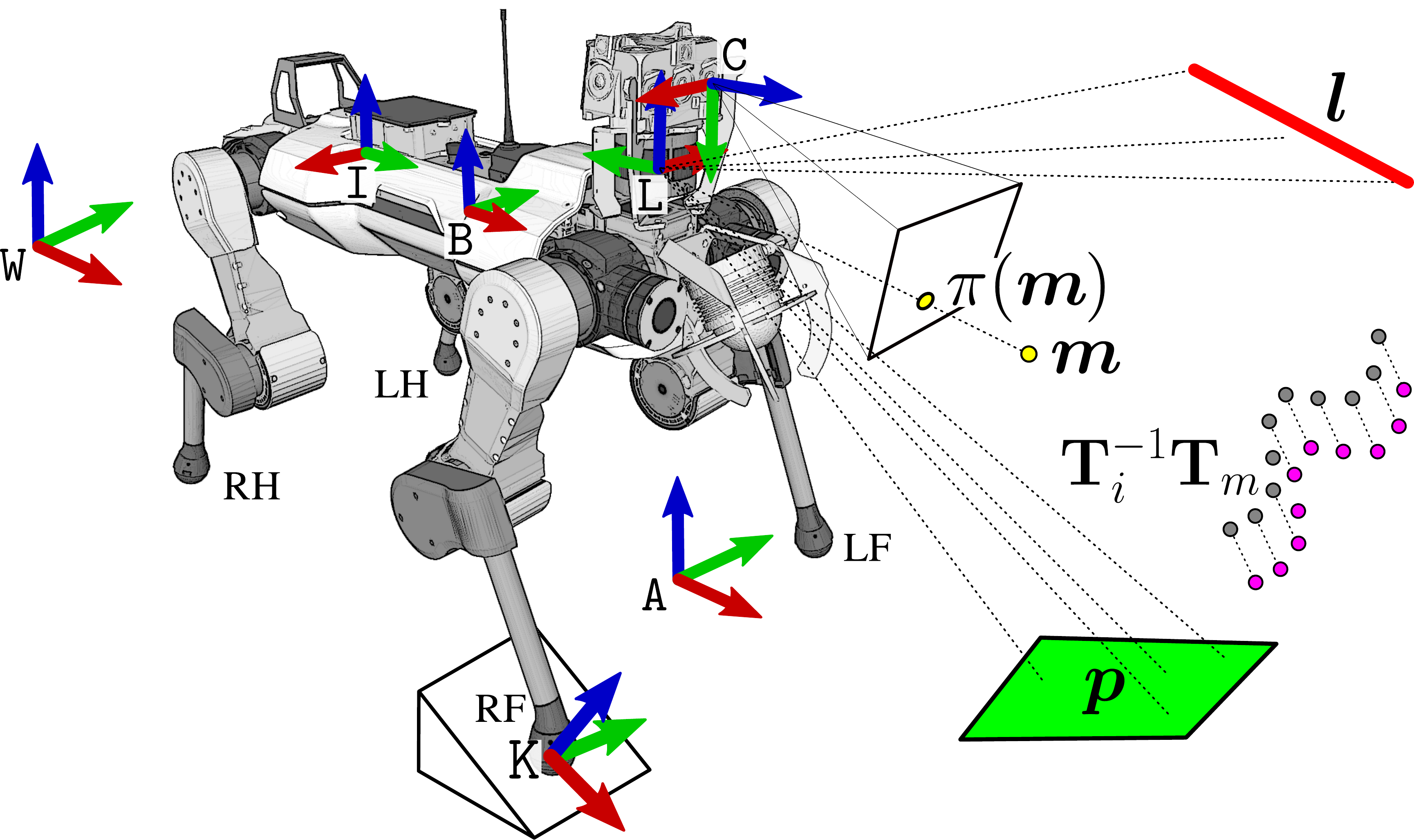}
\caption{Reference frames and landmark conventions. The world frame $\World$ is
fixed to Earth. The base frame $\Base$, the camera's optical frame $\Camera$,
the lidar frame $\Lidar$, and the IMU frame $\Imu$ are all rigidly attached to
the robot's chassis. The feet are conventionally named: Left Front (LF), Right
Front (RF), Left Hind (LH), and Right Hind (RH). When a foot touches the ground
(\eg RF), a contact frame $\Contact$ (perpendicular to the ground and parallel
to $\World$'s $y$-axis) is defined. The primitives tracked by the system are
points $\Landmark$, lines $\Line$, and planes $\Plane$. To improve numerical
stability, when a new plane feature is detected, an additional local fixed
anchor frame $\Anchor$ is defined. Finally, a relative pose factor (between
times $t_i$ and $t_m$ in figure) is created using lidar registration.}
\label{fig:coordinate-frames}
\end{figure}

\subsection{Notation}
Scalars are lowercase italics ($a, b, c, \dots$), matrices  are uppercase Roman
bold ($\Amat, \Bmat, \Cmat, \dots $), and vectors are lowercase Roman bold
($\mathbf{a}, \mathbf{b}, \mathbf{c}, \dots$). Reference frames are indicated in
typeface ($\mathtt{A},\mathtt{B},\mathtt{C}, \dots$) and for physical quantities
we follow the frame decorator rules from \cite{furgale2014notation}. States and
landmarks are bold italics ($\boldsymbol{a}, \boldsymbol{b}, \boldsymbol{c},
\dots$), while sensor measurements are uppercase calligraphics ($\calA, \calB,
\calC, \dots$). Finally, sets of time indices for measurements are uppercase
sans ($\mathsf{A}, \mathsf{B}, \mathsf{C}, \dots$). Where appropriate, a
time-dependent quantity will be shortened as $a(t_i) = a_i$.

\subsection{Frame Definitions}
The relevant reference frames are specified in \Figure
\ref{fig:coordinate-frames} and include: the fixed-world frame $\World$, the
base frame $\Base$, the IMU frame $\Imu$, the camera frame $\Camera$ (left
camera, when stereo), and the lidar frame $\Lidar$. When a foot is in contact
with the ground, a contact frame $\Contact$, fixed to Earth, is defined.
Finally, a local anchor frame $\Anchor$, also fixed to Earth, is defined for
lidar feature tracking, as detailed in Section \ref{sec:lidar-factors}.

\subsection{State Definition}
\noindent The robot state at time $t_i$ is defined as follows:
\begin{equation}
\State_i \triangleq \left[\R_i,\tran_i,\vel_i,\bg_i,\ba_i, \bw_i, \bv_i
\right] \in \SOthree \times \Real^{15} \label{eq:state-definition}
\end{equation}
where: $\R_i \in \SOthree$ is the orientation; $\tran_i \in \Realthree$ is the
position; $\vel_i\in \Realthree$ is the linear velocity; $\bg_i, \ba_i\in
\Realthree$ are the usual IMU gyroscope and accelerometer biases. We expand the
state with angular and linear velocity biases $\bw_i, \bv_i\in \Realthree$  to
model slippage, deformations, and other kinematics inaccuracies at the contact
point.

Unless otherwise specified, the position
$\tensor[_\world]{\tran}{_{\world\base}}$ and orientation $\R_{\world\base}$ of
the base are expressed in world coordinates, velocities of the base
$\tensor[_\base]{\vel}{_{\world\base}},
\tensor[_\base]{\rotvel}{_{\world\base}}$ are in base coordinates, IMU biases
$\tensor[_\imu]{\bias}{^{g}},\;\tensor[_\imu]{\bias}{^{a}}$ are expressed in the
IMU frame, and the velocity biases are expressed in the base frame,
$\tensor[_\base]{\bias}{_{\world\base}}{^\omega},
\;\tensor[_\base]{\bias}{_{\world\base}}{^v}$.

In addition to the robot state, we also estimate the position of visual and
lidar landmarks. Visual landmarks $\Landmark$ are parametrized as 3D points in
Euclidean space and projected onto the image plane via the function $\pi(\cdot)$
(yellow dot in \Figure \ref{fig:coordinate-frames}). Lidar landmarks are
parametrized as planar and linear geometric primitives (green plane $\Plane$ and
red line $\Line$ in \Figure \ref{fig:coordinate-frames}), as detailed in
\cite{Wisth2020lidar}. For brevity, we will later refer to any of the above
landmarks with $\boldsymbol{f}$.

We define the history of states and landmarks, visible up to the current time
$t_k$, as: \begin{equation} \States_k \defeq \lbrace \State_{i},
\Landmark_{\ell}, \Plane_{\ell}, \Line_{\ell} \rbrace_{i \in \mathsf{K}_k, \ell
\in \mathsf{F}_k} \end{equation} where $\mathsf{K}_k$ and $\mathsf{F}_k$ are the
sets of all keyframe and landmark indices, respectively.

\subsection{Measurements Definition}
We denote with $\calI_{ij}$ the IMU measurements received between two
consecutive keyframes $i$ and $j$. Each measurement includes the proper
acceleration $\tilde{\mathbf{a}}$ and the rotational velocity
$\tilde{\boldsymbol{\omega}}$, both expressed in the IMU frame. Similarly, we
define the kinematics measurements $\calK_{ij}$ which include the joint
positions $\tjp$ and velocities $\tjv$. The (mono or stereo) camera images and
lidar point clouds collected at time $t_i$  are expressed with $\calC_i$ and
$\calL_i$, respectively. In practice, camera and lidar measurements are received
at different times and frequencies, so they are first synchronized before being
integrated into the graph (see Section \ref{subsec:sensor-sync}).

The full set of measurements within the smoothing window is defined as:
\begin{equation}
\Measurements_k \defeq \lbrace
   \calI_{ij},  \calK_{ij}, \calC_{i}, \calL_{i}
\rbrace_{i, j \in \mathsf{K}_k}
\end{equation}

\subsection{Maximum-A-Posteriori Estimation}
We wish to maximize the likelihood of the measurements $\Measurements_k$
given the history of states and landmarks $\States_k$:
\begin{equation}
\States^\star_k = \argmax_{\States_k} p(\States_k|\Measurements_k) \propto
p(\States_0)p(\Measurements_k|\States_k)
\label{eq:posterior}
\end{equation}
Given that the measurements are assumed to be conditionally independent and
corrupted by white Gaussian noise, \Equation \eqref{eq:posterior} can
be formulated as a least squares minimization problem of the following form:
\begin{multline}
\States^{\star}_k = \argmin_{\States_k}
\|\residual_0\|^2_{\Sigma_0} +
\sum_{i \in \mathsf{K}_k} \bigg(
\|\residual_{\calI_{ij}}\|^2_{\Sigma_{\calI_{ij}}} +
\| \residual_{\calK_{ij}} \|^2_{\Sigma_{\calK_{ij}}} + \\
\|\residual_{\bias_{ij}}\|^2_{\Sigma_{\bias_{ij}}} + \|
\residual_{\calL_{i}} \|^2_{\Sigma_{\calL_i}}
+ \sum_{\ell \in \mathsf{F}_i} \| \residual_{\State_i, \boldsymbol{f}_\ell}
\|^2_{\Sigma_{\State_i, \boldsymbol{f}_\ell} }
\bigg)
\end{multline}
where each term is the squared residual error associated to a factor type,
weighted by the inverse of its covariance matrix, which will be detailed in
Section \ref{sec:twist-factors}. All the residuals (except the state prior
$\residual_0$) are added whenever a new keyframe $i$ is created. These include:
preintegrated IMU $\residual_{\calI_{ij}}$ and kinematic velocity
$\residual_{\calK_{ij}}$,  IMU and kinematic velocity biases
$\residual_{\bias_{ij}}$, lidar odometry $\residual_{\calL_{i}}$, and landmark
primitives $\residual_{\State_i, \boldsymbol{f}_\ell}$. The latter are further
divided into point, line, and plane residuals:
\begin{multline}
\sum_{\ell \in \mathsf{F}_i} \| \residual_{\State_i, \boldsymbol{f}_\ell}
\|^2_{\Sigma_{\State_i, \boldsymbol{f}_\ell} } =
\sum_{\ell \in \mathsf{M}_i} \| \residual_{\State_i, \Landmark_\ell}
\|^2_{\Sigma_{\State_i, \Landmark_\ell} } + \\
\sum_{\ell \in \mathsf{L}_i} \| \residual_{\State_i, \Line_\ell}
\|^2_{\Sigma_{\State_i, \Line_\ell} }
+ \sum_{\ell \in \mathsf{P}_i} \| \residual_{\State_i, \Plane_\ell}
\|^2_{\Sigma_{\State_i, \Plane_\ell} }
\end{multline}

\begin{figure}
 \centering
 \includegraphics[width=1.0\columnwidth]{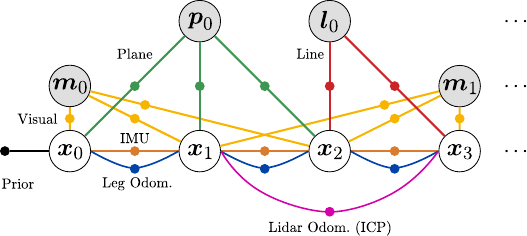}
 \caption{VILENS factor graph structure. The factors are: prior
(black), visual (yellow), lidar planes (green), lidar lines (red), preintegrated
IMU (orange), preintegrated velocity (from leg kinematics, blue), and lidar
odometry (from ICP registration, magenta). State nodes are white, while
landmarks are grey.}
\label{fig:factor-graph}
\end{figure}

\section{Factor Graph Formulation}
\label{sec:twist-factors}
In this section, we describe the measurements, residuals and covariances of the
factors which form the graph shown in \Figure \ref{fig:factor-graph}. For
convenience, we summarize the IMU factors from \cite{Forster2017} in Section
\ref{sec:imu-residuals}; our novel velocity factor is detailed in Section
\ref{sec:velocity-factor}; Sections \ref{sec:bias-residuals} and
\ref{sec:visual-factors} describe the bias and stereo visual residuals, which
are adapted from \cite{Forster2017, Wisth2020icra} to include the velocity bias
term and support for fisheye cameras, respectively. We also briefly introduce
the lidar factor residuals from \cite{Wisth2020lidar} in Section
\ref{sec:lidar-factors}. Finally, the novel lidar registration residual is
detailed in
Section \ref{sec:icp-factors}.

\subsection{Preintegrated IMU Factors}
\label{sec:imu-residuals}
As in \cite{Forster2017}, the IMU measurements are preintegrated to constrain
the pose and velocity between two consecutive nodes of the graph, as well as
provide high-frequency state updates between them. This uses a residual of the
form:
\begin{equation}
	\residual_{\calI_{ij}}  = \left[ \residual^\transpose_{\Delta
		\R_{ij}}, \residual^\transpose_{\Delta \vel_{ij}},
	\residual^\transpose_{\Delta \tran_{ij}} \right]
\end{equation}
where the individual elements of the residual are defined as:
\begin{align}
	\residual_{\Delta \R_{ij}}   &= \Log{
	\Delta\tilde{\R}_{ij}(\bg_{i}) }
	\R^\transpose_{i} \R_j \label{eq:rot-residual}\\
	\residual_{\Delta \vel_{ij}}  &=   \R^\transpose_{i} \left( \vel_j
	-
	\vel_{i} - \gravity \Delta t_{ij} \right) - \Delta
	\tilde{\vel}_{ij}(\bg_{i}, \ba_{i}) \\
	\residual_{\Delta \tran_{ij}} &=  \R^\transpose_{i} \left( \tran_j
	-
	\tran_{i} - \vel_{i}\Delta t_{ij} - \frac{1}{2}\gravity \Delta
	t_{ij}^2 \right) \nonumber\\
	&\hspace{3cm}- \Delta \tilde{\tran}_{ij}(\bg_{i}, \ba_{i})
\label{eq:imu-residual}
\end{align}
As explained in more detail in \cite{Forster2017}, this formulation does not
require recomputation of the integration between the two keyframes connected by
the factor every time the linearization point changes. This makes the fusion of
the high-frequency signal from the IMU and lower frequency measurements from
camera and lidar computationally feasible.

The \Equations \eqref{eq:rot-residual} -- \eqref{eq:imu-residual} depend on the
optimized states and preintegrated measurements $\dtRij, \dtvij,
\preintPmeas_{ij}$. For convenience, we report them here (for the incorporation
of the bias update and other details, see \cite{Forster2017}):
\begin{align}
 \preintRmeas_{ij} &= \prod_{k=i}^{j-1} \expmap\left(\left(
\tilde\rotvel_k - \bias^g_i \right) \Delta t\right)
\label{eq:preint-imu-rot-meas}\\
\preintVmeas_{ij}&= \sum_{k=i}^{j-1} \! \preintRmeas_{ik}  \! \left(
\tilde\acc_k \!-\! \bias^a_i \right)  \Delta t \\
\preintPmeas_{ij} &= \sum_{k=i}^{j-1}
      \Big[
        \preintVmeas_{ik}\Delta t
        \!+\! \frac{1}{2} \preintRmeas_{ik}
\left(\tilde\acc_k \!-\! \bias^a_i\right)\Delta t^2  \Big]
\end{align}
These are associated to the corresponding preintegrated noises, which can be
expressed in iterative form:
\begin{multline}
 \begin{bmatrix}
\dPhi_{i,k+1} \\
\velpert_{i,k+1} \\
\tranpert_{i,k+1}
 \end{bmatrix} = \begin{bmatrix}
\preintRmeas_{k,k+1}^\transpose & \Zero & \Zero \\
-\preintRmeas_{ik}(\tacc_k - \ba_i)^\wedge \dt & \eye & \Zero \\
-\frac{1}{2}\preintRmeas_{ik}(\tacc_k - \ba_i)^\wedge \dt^2 & \eye\dt & \eye
 \end{bmatrix} \\ \begin{bmatrix}
\dPhi_{i,k} \\
\velpert_{i,k} \\
\tranpert_{i,k}
 \end{bmatrix} + \begin{bmatrix}
 \JacR^k \dt & \Zero \\
 \Zero & \preintRmeas_{ik}\dt \\
 \Zero & \frac{1}{2} \preintRmeas_{ik} \dt^2
 \end{bmatrix} \begin{bmatrix}
 \etag \\
 \noise^a
 \end{bmatrix}\label{eq:preint-imu-rot-noise}
\end{multline}
from which the covariance $\Cov_{\measimu_{ij}}$ can be computed.

\subsection{Preintegrated Leg Odometry Factors}
\label{sec:velocity-factor}
Inspired by the preintegrated IMU factor, we define here the preintegrated
velocity factor for leg odometry. This factor is used to estimate the position
of the robot from  high-frequency joint kinematics measurements.

To derive a velocity signal from leg kinematics, we first need to estimate the
stance legs, \ie the set of legs in stable contact with the ground (Section
\ref{sec:stance-estimation}). Then, we fuse the odometry from these legs as a
combined velocity measurement (Section \ref{sec:leg-odometry}). To model the
nonlinearities at the contact point, the linear velocity bias and its residuals
are introduced in Section \ref{sec:velocity-bias}. The preintegrated
measurements and their residuals are described in Sections
\ref{sec:preintegrated-measurements} and \ref{sec:velocity-residuals},
respectively. Finally, a special case of this twist bias is outlined in Section
\ref{sec:special-cases}.

\subsubsection{Stance Estimation}
\label{sec:stance-estimation}
The feet of the robot are approximated as points. For simplicity, the contact
point is assumed to be on a fixed point at the center of the foot of the robot,
which in our case is a soft rubber sphere with a \SI{2}{\centi\meter} radius.
Since no direct force measurement is assumed to be available, we compute the
Ground Reaction Force (GRF) at each foot from the dynamics equation of motion:
\begin{equation}
\Force = - \left( \Jacp(\jp)^\transpose \right)^{\dagger}
\left( \Torque + \mathbf{h}(\jp) + F^\transpose
\begin{bmatrix} \dot{\rotvel} \\ \dot{\vel} \end{bmatrix}
\right)
\end{equation}
where $^\dagger$ indicates the Moore-Penrose pseudo inverse operation, $\jp$ are
the joint positions, $\Jacp(\cdot) \in \Real^{3 \times 3}$ is the Jacobian of
the forward kinematics function relative to the point foot (later indicated with
$\fkp(\cdot) \in \Real^3$), $\Torque$ are the joint torques, $\mathbf{h}(\jp)$
are the Coriolis terms, and $F$ is the matrix of spatial forces required at the
floating base to support unit accelerations about each joint variable
\cite{roybook}. In practice, both $\dot{\rotvel}$ and $\dot{\vel}$ are obtained
from the IMU and projected into the base frame, where the angular acceleration
is computed by numerical differentiation and the linear velocity is computed by
compensating for the gravity using the latest orientation estimate.

For each leg that is in contact with the ground, we assume \emph{rigid,
nonslipping contact}. By thresholding the vertical component of the GRF for each
foot, we get the set of binary contact states for all the legs. We indicate the
subset of the legs in contact as $S \subseteq L$, with $L = \lbrace
\text{LF},\text{RF},\text{LH},\text{RH} \rbrace$ (\Figure
\ref{fig:coordinate-frames}).

\subsubsection{Combined Kinematic Velocity Measurement}
\label{sec:leg-odometry}
Given a leg $s \in S$, we can compute the linear velocity of the robot's
floating base at time $t_i$ as follows:
\begin{equation}
\vel_{s} = -\Jacp(\jp)\jv - \wb \times \fkp(\jp)
\label{eq:no-slip}
\end{equation}
Both the joint positions and velocities are measured from encoders and
corrupted by additive zero-mean Gaussian noise:
\begin{align}
\tilde{\jp} &= \jp + \etajp \label{eq:joint-pos-noise}\\
\tilde{\jv} &= \jv + \etajv \label{eq:joint-vel-noise}
\end{align}
where $\etajp, \etajv$ are available from the sensor specifications. After
substituting \Equations \eqref{eq:joint-pos-noise} and
\eqref{eq:joint-vel-noise} into
\Equation \eqref{eq:no-slip}, we can formulate a linear velocity measurement
\cite{Bloesch2013}:
\begin{equation}
\vel_{s} =
- \Jacp(\tilde{\jp}-\etajp)\cdot(\tilde{\jv} - \etajv)
- \rotvel \times \fkp(\tilde{\jp}-\etajp)
\label{eq:legodo-single-leg}
\end{equation}
As detailed in Appendix \ref{sec:derivation-leg-odo-velocity}, the
noise terms from \Equation \eqref{eq:legodo-single-leg} can be separated, so
the measurement can be expressed as:
\begin{align}
\tilde{\vel}_{s} &=
\vel_{s} + \etav_s  \label{eq:legodo-with-noise} \\
\etav_s &= - \left(\hessian{\tjp}\tjv+\rotvel^\wedge\Jacpp{\tjp}\right)\etajp
-\Jacpp{\tjp} \etajv \label{eq:etav}
\end{align}
where $\hessian{\cdot} \in \Real^{3 \times 3 \times 3}$ is the
Hessian of the forward kinematics function, $\fkp(\jp)$.

Equation \eqref{eq:legodo-with-noise} is valid only when the leg $s$ is in
contact with the ground. Since foot contacts happen intermittently while the
robot walks, multiple legs can be in contact simultaneously. As each velocity
measurement from \Equation \eqref{eq:legodo-with-noise} is associated with a
Gaussian noise term $\etav_s$, it would be possible to add to the graph one
independent velocity measurement per stance leg. This would, however,
unnecessarily increase the graph complexity, as a closed form for fusing the
measurements from the stance legs into one can be computed instead
\cite{barfoot2014tro}. Additionally, treating the leg velocities separately
would require explicit handling of contact switching every time a new step is
made to ensure all measurements are used \cite{Hartley2018b}.

From \Equation \eqref{eq:etav}, since $\etav_s$ is a linear combination of
zero-mean Gaussians, it is also a zero-mean Gaussian with covariance $\Cov_s$.
Therefore, a combined velocity measurement for all the stance legs can then be
computed as a weighted average of the velocity measurements of each stance leg,
based on the information matrix $\InfoMat_s = \Cov_s^{-1}$:
\begin{align}
 \Cov_v &= \left(\sum_{s \in S} \InfoMat_s\right)^{-1} \\
 \tilde{\vel} &= \Cov_v \sum_{s \in
S}
\left(\InfoMat_s \tilde{\vel}_s \right) \\
\etav &\sim \gaussian{\boldsymbol{0}}{\Cov_v}
\label{eq:legodo-combined}
\end{align}
where the compound velocity measurement noise $\etav$ is sampled from a
zero-mean Gaussian with covariance $\Cov_v$.

We now have a linear velocity measurement\footnote{Not to be confused with the
IMU preintegrated velocity measurement, $\preintVmeas$.} $\tv$ and its noise
$\etav$ that can be used for the preintegrated velocity factor:
\begin{equation}
 \tv = \vel + \etav
 \label{eq:vel-meas-final}
\end{equation}

\subsubsection{Velocity Bias}
\label{sec:velocity-bias}
On slippery or deformable terrains, the constraint from \Equation
\eqref{eq:no-slip} will not be true, resulting in incorrect leg velocities, and
thus drift in the final odometry estimate. In our experience, this velocity
drift is locally constant and is gait- and terrain-dependent (see Sections
\ref{subsection:Motivation} and \ref{sec:special-cases}, and \Figures
\ref{fig:position-drift-motivation} and \ref{fig:foot-contact-sequence}).

For these reasons, we relax \Equation \eqref{eq:no-slip} by adding a
slowly varying bias term $\bv$ to \Equation \eqref{eq:vel-meas-final}:
\begin{equation}
\tilde{\vel} = \vel + \bv + \etav
\label{eq:vel-measurement}
\end{equation}
This term incorporates the characteristic drift caused by leg or terrain
compression, slippage, and impacts occurring at the contact point.

\subsubsection{Preintegrated Velocity Measurements}
\label{sec:preintegrated-measurements}
We derive the preintegrated linear velocity and noise only, as the preintegrated
rotation measurement $\Delta\tilde{\mathbf{\Theta}}_{ij}$ and noise
$\delta\boldsymbol{\theta}_{ij}$ have the same form as \Equations
\eqref{eq:preint-imu-rot-meas} and \eqref{eq:preint-imu-rot-noise}:
\begin{align}
  \Delta\tilde{\mathbf{\Theta}}_{ij} &= \prod_{k=i}^{j-1} \expmap\left(\left(
\tilde\rotvel_k - \bias^\w_i \right) \Delta t\right)
\label{eq:preint-imu-rot-meas-twist}\\
\delta\boldsymbol{\theta}_{ij} &= \sum_{k=i}^{j-1}
\Delta\tilde{\mathbf{\Theta}}^\transpose_{k+1,j}\JacR^k\etaw\dt
\label{eq:preint-imu-rot-noise-twist}
\end{align}
The position at time $t_j = t_i + \dt$ is:
\begin{equation}
\tran(t_j) = \tran(t_i) + \int_{t_i}^{t_j} \vel(\tau)\; \mathrm{d}\tau
\end{equation}
Assuming constant velocity between $t_i$ and $t_j$, we can iteratively calculate
the position in discrete time domain form\footnote{For simplicity, we keep the
symbol $\etav$ for the noise in the discrete domain.}:
\begin{equation}
\tran_j  =  \tran_i + \sumjmo{\R_{k} ( \tvk - \bvi - \etav_k) \dt}
\label{eq:iterative-pos}
\end{equation}
From \Equation \eqref{eq:iterative-pos} a relative measurement can be
obtained\footnote{The variable name change is to avoid confusion with the
position measurements and noise from the IMU factor.}:
\begin{equation}
\dkij = \R_i^\transpose(\tran_j-\tran_i) =
\sumjmo{\dR_{ik}(\tvk-\bv_i-\etav_k)\dt)}
\label{eq:rel-meas}
\end{equation}
With the substitution $\dRik =
\dtTheta_{ik}\Exp{-\delta\boldsymbol{\theta}_{ik}}$ to include the preintegrated
rotation measurements (from \Equations \eqref{eq:preint-imu-rot-meas-twist} --
\eqref{eq:preint-imu-rot-noise-twist}), and the approximation $\Exp{\rotvec}
\simeq \Identity + \rotvec^{\wedge}$, \Equation \eqref{eq:rel-meas} becomes:
\begin{equation}
\dkij \simeq \sumjmo{ \dtTheta_{ik} (I -\dVtheta_{ik}^\wedge)(\tvk - \bvi -
\etav_k)
\dt }
\label{eq:rel-meas-2}
\end{equation}
By separating the measurement and noise components of \Equation
\eqref{eq:rel-meas-2} and ignoring higher order terms, we can define the
\textit{preintegrated  leg odometry position measurement} $\dtpij$ and
\textit{noise} $\depij$ as:
\begin{align}
\dtpij  &\defeq  \sumjmo{ \dtTheta_{ik} (\tvk - \bvi)\dt }
\label{eq:rel-meas-3}\\
 \depij  &\defeq  \sumjmo{\dtTheta_{ik} \etav_k \dt - \dtTheta_{ik} (\tvk -
\bvi)^\wedge
 \dVtheta_{ik} \dt}
 \label{eq:preint-noise-model}
\end{align}
Note that both quantities still depend on the twist biases $\bw, \bv$. When
these change, we would like to avoid the recomputation of \Equation
\eqref{eq:rel-meas-3}. Given a small change $\delta \bias$ such that $\bias =
\bar{\bias} + \delta\bias$, we use a first order approximation to find the new
measurement, as done in \cite{Forster2017}:
\begin{multline}
\dtpij(\bw,\bv) \simeq
\dtpij(\bar{\bias}^\omega,\bar{\bias}^v)
+ \\
\frac{\partial\dtpij}{\partial\bw}\delta\bw
+ \frac{\partial\dtpij}{\partial\bv}\delta\bv
\end{multline}

\subsubsection{Residuals}
\label{sec:velocity-residuals}
The factor residuals include rotation and translation:
\begin{align}
\residual_{\calK_{ij}}  &= \left[ \residual^\transpose_{\Delta
		\tilde{\mathbf{\Theta}}_{ij}},\; \residual^\transpose_{\dkij} \right] \\
	\residual_{\Delta \mathbf{\Theta}_{ij}}   &= \Log{
	\Delta\tilde{\mathbf{\Theta}}_{ij}(\bw_{i}) }
	\R^\transpose_{i} \R_j \label{eq:rot-residual-twist}\\
\residual_{\dkij} &= \R_i^\transpose \left( \tran_j - \tran_i \right)
-
\dtpij(\bw_i,\bvi)
\end{align}
Since \Equation \eqref{eq:rot-residual-twist} has the same form as \Equation
\eqref{eq:rot-residual}, when the same angular velocity measurements are used
for both the IMU and leg odometry factors, the following relations hold:
\begin{equation}
 \Delta\tilde{\mathbf{\Theta}}_{ij} = \preintRmeas_{ij} \quad\quad
 \bw = \bg \quad\quad
 \dVtheta_{ij} = \dPhiij
\end{equation}
In this case, we can avoid double counting the IMU signal by setting the
rotational residual $\residual_{\Delta \mathbf{\Theta}_{ij}}$ to zero. This was
not possible in \cite{Wisth2020icra}, where an external filter was used.

\subsubsection{Covariance}
After simple manipulation of \Equation \eqref{eq:preint-noise-model}, the
covariance of the residual $\residual_{\calK_{ij}}$ can be computed iteratively:
\begin{equation}
\Cov_{i,k+1}^\calK = \Amat \Cov_{i,k}^\calK \Amat^\transpose +
\Bmat \Cov_{\noise}^{\calK} \Bmat^\transpose
\end{equation}
where the first term evolves from an initial condition of $\Cov^{\calK}_{i,i} =
\Zero$, while the second term $\Cov_{\noise}^{\calK}$ is fixed and taken from
sensor specifications. The complete derivation of the multiplicative terms
$\Amat$ and $\Bmat$ are detailed in Appendix \ref{sec:iter-noise-propagation}.

\subsection{Bias Residuals}
\label{sec:bias-residuals}
The bias terms are intended to change slowly and are therefore modeled as a
Gaussian random walk. The residual term for the cost function is as follows:
\begin{multline}
\|\residual_{\bias_{ij}}\|^2_{{\Sigma}_{\bias_{ij}}} \defeq
  \|{\bg_j}-{\bg_i}\|^2_{\Sigma_{\bg_{ij}}} + \\
\|{\ba_j}-{\ba_i}\|^2_{\Sigma_{\ba_{ij}}}
+
\|{\bv_j}-{\bv_i}\|^2_{\Sigma_{\bv_{ij}}}
\end{multline}
where the covariance matrices are determined by the expected rate of change of
these quantities. In particular, $\Cov_{\bg_{ij}}$, $\Cov_{\ba_{ij}}$ are
available from IMU specifications, while $\Cov_{\bv_{ij}}$ depends on the drift
rate of the leg odometry bias, which is found empirically.

\subsection{Visual Factors}
\label{sec:visual-factors}
We use two main factors related to visual measurements. The first is a
traditional reprojection error given by \cite{Wisth2019}:
\begin{equation}
\residual_{\State_i,\Landmark_\ell} =
\left( \begin{array}{c}
\pi_u^L(\R_i,\tran_i, \Landmark_\ell) - u^L_{i,\ell} \\
\pi_u^R(\R_i,\tran_i, \Landmark_\ell) - u^R_{i,\ell} \\
\pi_v(\R_i,\tran_i, \Landmark_\ell) - v_{i,\ell}
\end{array} \right)
\label{eq:visual-residual}
\end{equation}
where $(u^{L}, v), (u^{R}, v)$ are the pixel locations of the detected landmark.
If only a monocular camera is available then the second element of \Equation
\eqref{eq:visual-residual} is not used.

The second factor uses the overlapping fields of view of the lidar and camera
sensors (where applicable) to provide depth estimates for visual features, as
described in \cite{Wisth2020lidar}.

In addition to the standard camera projective model, in this paper we also
introduce support for fisheye cameras with equidistant distortion
\cite{KannalaBrandt2006}. As demonstrated in \cite{largefovbenefit}, a high
camera FoV allows for tracking
of features for longer periods of time, but for large open areas the loss of
pixel density becomes significant for a diagonal FoV of \SI{180}{\degree} or
more. To maximize versatility in underground scenarios, our fisheye
configurations have a moderately large diagonal FoV of
\SIrange{150}{165}{\degree} (Table \ref{tab:specs}).

To add the landmark to the
graph, we first detect and
track features in the original, distorted image to avoid costly image
undistortion. We then undistort the individual feature locations before adding
them to the factor graph using \Equation \eqref{eq:visual-residual}.

\begin{figure}
\centering
\includegraphics[width=0.49\columnwidth]{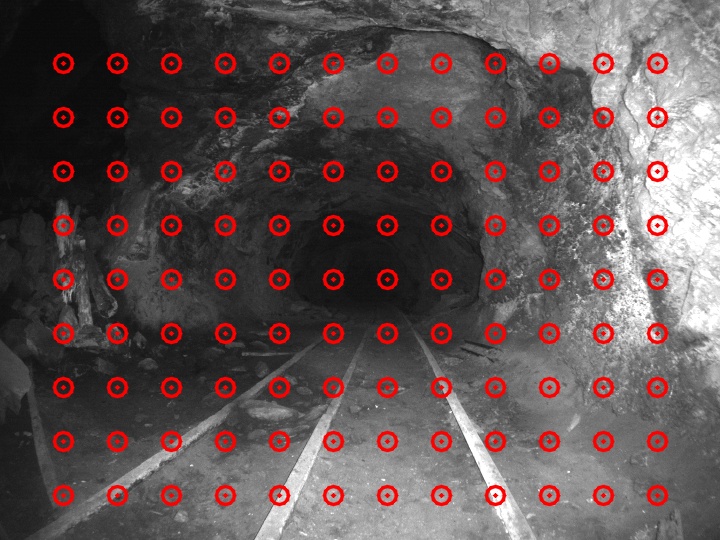}%
\hspace{0.004\columnwidth}%
\includegraphics[width=0.49\columnwidth]{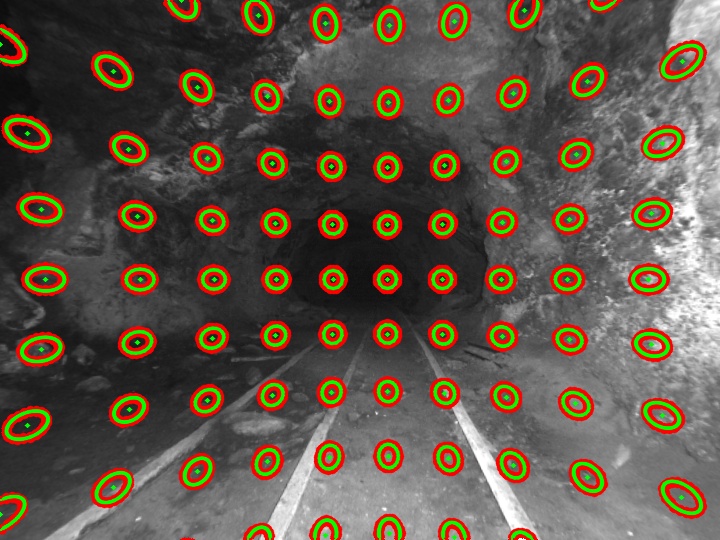}%
\caption{Example of fisheye covariance warping and fitting. \textit{Left:}
Original samples (red points) and 1-$\sigma$ bound of 9 pixels (red circles) in
the distorted image. \textit{Right:} the undistorted points from the 1-$\sigma$
bound (red) and their ellipse fit (green) in the undistorted image. Notice that
the undistorted covariance becomes larger near the edge of the image where there
is greater distortion.}
\label{fig:fisheye-covariance}
\end{figure}

As shown in \Figure \ref{fig:fisheye-covariance}, the distortion far from the
focal point can be quite large. This means the angular resolution of the camera
varies across the image. To correctly model the visual feature uncertainty,
$\Cov_{\State_i, \Landmark_\ell}$, we first select a set of points around the
landmark in the distorted image, based on the specified visual feature tracker
uncertainty. We then undistort these points and use a least-squares method to
fit an ellipse to these points. For ellipse fitting we use at least \num{6}
points since the covariance ellipse is, in general, is not aligned to the $x$ or
$y$ axes. This gives us an accurate uncertainty matrix in the undistorted image
coordinates. Since the distortion is constant, the undistortion map and
covariance ellipses can be precomputed for efficiency, and thus have no impact
on runtime performance.

We use $\Sigma_{\State_i, \Landmark_\ell}$ of between \num{1} and \SI{2}{pixels}
(adapted by the fisheye covariance warping if applicable).

\subsection{Plane and Line Factors}
\label{sec:lidar-factors}
We extract and track geometric primitives, specifically planes and lines, over
successive lidar scans. This is similar to how visual features are tracked.
Planes are defined using the Hessian norm form \cite{Wisth2020lidar}:
\begin{equation}
\Plane = \left\{\langle \hat{\mathbf{n}}, d \rangle \in
\mathbb{R}^{4} ~\vert~ \hat{\mathbf{n}}\cdot (x,y,z) + d = 0 \right\}
\end{equation}
while lines use the minimal parametrization from \cite{minlie}:
\begin{equation}
\Line = \left\{\langle \R,(a, b)\rangle \in \SOthree \times
\mathbb{R}^{2}\right\}
\end{equation}
In this paper, we extend the plane and line factors from \cite{Wisth2020lidar}
to support  local linearization points. This is important in large scale
environments, where the transformation of the plane from world to sensor
coordinates can cause numerical instability in the optimization (small changes
in the angle of the normal can cause very large changes in the position of the
plane or line). This instability increases as the sensor gets further from the
origin of the world frame.

Therefore, inspired by \cite{Kaess2015}, we introduce a local linearization
frame for planes and lines which we call the anchor frame, $\Anchor$. This both
decreases convergence time \cite{Kaess2015} and increases numerical stability.
$\Anchor$ is arbitrarily defined at the first frame in which the landmark is
observed.

When a plane $\tilde{\Plane}_\ell$ is measured at time $t_i$, the corresponding
residual is the difference between $\tilde{\Plane}$ and the estimated plane
$\Plane_\ell$ is transformed into the local reference frame:
\begin{equation}
\residual_{\State_i, \Plane_\ell} = \left(
\left(\T^{-1}_{\World\Anchor}
\T_{\World\Base} \right) \otimes \Plane_\ell
\right) \ominus \tilde{\Plane}_\ell \label{eq:plane-residual2}
\end{equation}
where $\T_{\World\Anchor}$ is the pose of the robot at the time where the plane
is first detected, $\T_{\World\Base} = [\tran_i, \R_i]$ is the current pose
estimate of the robot.

Similarly, the residual between a measured line $\tilde{\Line}_i$ and its
prediction is defined as follows:
\begin{equation}
\mathbf{r}_{\State_i, \Line_\ell} = \left(\left(\T^{-1}_{\World\Anchor}
\T_{\World\Base} \right) \boxtimes \Line_\ell\right)
\boxminus \tilde{\Line}_\ell \label{eq:line-residual}
\end{equation}
where $\otimes$, $\boxtimes$ apply a transformation and $\ominus$, $\boxminus$
are difference operators to planes and lines, respectively
\cite{Wisth2020lidar}.

 The line and plane covariances $\Cov_{\State_i, \Line_\ell},\;\Cov_{\State_i,
\Plane_\ell}$ are
determined by analysing the covariance of the inlier points for each feature.
This results in typical covariances of \SI{10}{\centi\meter} and
\SI{3}{\deg}.

\subsection{Lidar Registration Factor}
\label{sec:icp-factors}
The lidar feature tracking allows for continuous motion estimation at the full
lidar frame rate. Approaches like \cite{liosam2020shan} instead accumulate the
features into a local submap for a certain number of frames before integrating
them into the factor graph. Local submapping allows for accurate pose estimation
refinement, but at a lower frequency. In this work, in addition to the feature
tracking, we also integrate fine-grained lidar registration with local
submapping.  The registration is based on the Iterative Closest Point
(ICP) approach by \etalcite{Pomerleau}{Pomerleau12libpointmatcher}. To make the
registration more robust, a local submap is maintained with
all the scans successfully registered in the past \SI{5}{\meter} travelled. The
submaps constitutes the reference cloud of the ICP registration process.

Although ICP typically runs at a lower frequency than the other signals, it is
able to take advantage of the high quality motion prior provided by the other
modules that run at higher frequency (\eg IMU) to motion correct scans.

ICP odometry measurements are added into the factor graph as relative pose
factors between nonconsecutive keyframes $i$ and $m$ (\eg nodes 1 and 3 in
\Figure \ref{fig:factor-graph}):
\begin{equation}
\residual_{\calL_{i}} = \Phi(\widetilde{\T}_i^{-1}
\widetilde{\T}_m, \T_i^{-1} \T_m)
\end{equation}
where $\T$ is the estimated pose, $\widetilde{\T}$ is the estimate from the ICP
module, and $\Phi$ is the lifting operator defined in \cite{Forster2017}. Note
that ICP registration is prone to failure in environments with degenerate
geometries, such as long tunnels. For this reason, a robust cost function is
used to reject the factor in such degenerate situations.

For simplicity, the lidar registration covariance $\Cov_{\calL}$ is set as a
constant value, determined empirically during preliminary experiments and left
unchanged for all our experiments, with potential outliers being handled by the
robust cost function.

\section{Implementation}
\label{sec:implementation}
\begin{figure}
\centering
\includegraphics[width=\columnwidth]{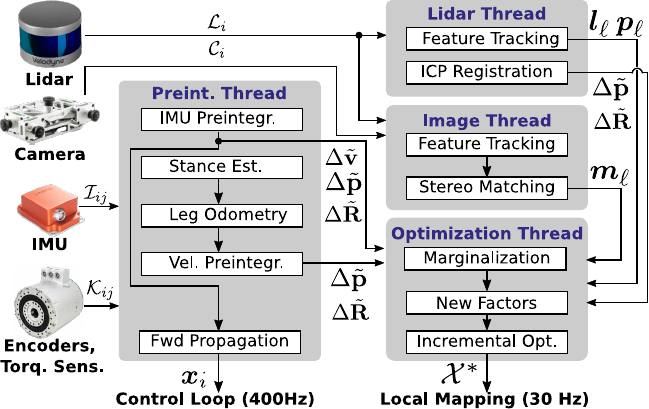}%
\caption{Block diagram of the VILENS algorithm. Three independent threads
(preintegration, camera, and lidar) process the data while a fourth thread
performs the optimization on the data already processed by the
other threads.}
\label{fig:state-estimation-architecture}
\end{figure}

The architecture of the VILENS state estimator is shown in \Figure
\ref{fig:state-estimation-architecture}. Three parallel threads process the
preintegration of IMU and kinematics, camera feature tracking, lidar processing
(both feature tracking and ICP registration) and a fourth carries out the
subsequent optimization. A forward propagated state from the IMU factor is
output from the preintegration thread at IMU frequency (\ie \SI{400}{\hertz}).
This is used to motion correct and synchronize the lidar point clouds  and is
available for high-frequency tasks such as control (for simplicity, we assume
$\calL_{i}$ is already undistorted in \Figure
\ref{fig:state-estimation-architecture}). When a new keyframe is processed, the
preintegrated measurements and tracked landmarks are collected by the
optimization thread, while the other threads process the next set of
measurements. When the optimization step is complete, the optimal set of states
is produced at the keyframe rate for use by local mapping and path planning.

The factor graph is solved using a fixed lag smoothing framework based on the
efficient incremental optimization solver iSAM2 \cite{Kaess2012}, implemented as
part of the GTSAM library \cite{Dellaert2017}. For these experiments, we use a
lag time of \SI{5}{\second}. All visual and lidar factors are added to the graph
using the Dynamic Covariance Scaling (DCS) \cite{MacTavish2015} robust cost
function to reduce the effect of outliers.

\subsection{Sensor Synchronization}
\label{subsec:sensor-sync}
To perform joint optimization across all the sensor modalities, the measurements
first need to be synchronized.  As in \cite{Wisth2019}, the IMU and kinematic
measurements, which have high frequency, are interpolated to match the camera
timestamps. To synchronize the lidar measurements with the camera, we instead
follow the approach described in \cite{Wisth2020lidar}. When a new point cloud
is received, its points are motion compensated using the IMU propagated state
(from the preintegrated IMU factor) using the closest camera keyframe timestamp
as reference, instead of the start or end of the timestamp, as commonly done. In
this way, the minimum number of nodes will be added to the graph. The
optimization is performed \emph{jointly} between IMU, kinematics, camera, and
lidar inputs. This also ensures a fixed output frequency, \ie the camera
keyframe rate.

\subsection{Forward Kinematics and Dynamics}
The forward kinematics and dynamics are implemented in RobCoGen
\cite{frigerio2016joser}, which is a computationally efficient kinematics and
dynamics solver, as demonstrated in this comparison paper \cite{neuman2019iros}.
RobCoGen takes the current joint configuration and base state to calculate the
position, velocity, and force at each of the feet. The covariance values for
the kinematics are taken directly from the encoder datasheet.

\subsection{Twist Bias Modes}
\label{sec:special-cases}
The introduction of a twist bias is motivated by non-ideal contact events which
can cause drift in the estimated velocity. Therefore, if no new contact events
are occurring (\ie the robot is not making any new steps), this velocity bias
should not be present.

To account for this, when no footsteps are being made we disable the twist bias
and simply use relative pose factors instead. This is important for legged
robots since the robot can still move (\eg roll and pitch) while the feet are
stationary. This is in contrast to the zero velocity mode presented in
\cite{Wisth2020icra} where access to the contact states was not possible due to
the use of an external filter to process the kinematics.

In practice, we detect this \textit{low drift} state when the majority of the
robot's feet (\ie 3 or 4 feet) are in constant contact with the ground for an
extended period of time (\ie more than \SI{200}{\milli\second}). As soon as new
footsteps are made we exit this low-drift state and return to normal operation.

\subsection{Visual and Point Cloud Feature Tracking}
We detect visual features using the FAST corner detector, and track them between
successive frames using the KLT feature tracker. Outliers are rejected using a
RANSAC-based fundamental matrix consistency check (similar to \cite{Qin2018}).
For point cloud features, we use the feature tracking approach based on
\cite{Wisth2020lidar} to extract and track geometric primitives (namely planes
$\Plane_{\ell}$ and lines $\Line_{\ell}$) over time.

Keyframes are added when mean visual feature movement between
consecutive frames is greater than a threshold (\SI{1}{pixel}), with minimum
(\SI{5}{\hertz}) and maximum (\SI{15}{\hertz}) keyframe frequency bounds.

\subsection{Zero Velocity Update Factors}
To limit drift and factor graph growth, we continuously query if the robot is
stationary by using a voting mechanism. If the majority of sensor modalities
detect no motion then we add a zero velocity constraint to the graph. This
method can detect when the robot is stationary, even when one or more sensor
inputs are not available (\eg legs are not in contact with the ground).

For example, the IMU and leg odometry threads report zero velocity when position
(rotation) is less than \SI{0.1}{\milli\meter} (\SI{0.5}{\degree}) between two
keyframes. The image thread reports zero velocity when average feature movement
between frames is less than \SI{0.5}{\text{pixels}}. The lidar thread reports
zero velocity when the motion induced by tracking planes and lines is less than
\SI{0.2}{\meter} and \SI{5}{\degree}, respectively.

\subsection{Calibration}
We use the open source camera and IMU calibration toolbox Kalibr
\cite{Furgale2013kalibr} to compute the intrinsic and extrinsic
calibrations of the cameras. The IMU, lidar, and kinematics positions are known 
from accurate
CAD models and are rigidly attached to the frame of the robot.

All sensors are hardware-synchronized where possible (using either Precision
Time Protocol (PTP), EtherCAT, or hardware triggering). The only exception is
the Realsense D435i cameras which are software-synchronized using the
manufacturer's driver, before removing any clock drift using continuous-time
maximum likelihood estimation \cite{Ramezani2020newer}.

\subsection{Initialization}
We initialize the system by averaging the first \SI{1}{\second} of IMU data at
system start-up (assuming the IMU is stationary). This allows us to find the
gyroscope bias and initial estimates for orientation of the gravity vector. As
they cannot be estimated from stationary data \cite{Martinelli2014},
accelerometer biases are initially assumed to be either zero or a known constant
value. The accelerometer biases would then converge to their true
value once the robot starts moving, thanks to the robot's 3D walking motion. The
scale is known from either stereo correspondence or lidar measurements.

\section{Experimental Results}
\label{sec:results}
In this section, we will describe the platforms and the dataset used in our
evaluations (Sections \ref{sec:platforms} and \ref{sec:datasets}). We will then
demonstrate how the tight integration of kinematics into a factor graph
framework allows for easy combination of the different sensor modalities
(Section \ref{sec:multisensor-results}). Finally, we will end with an analysis
of specific challenging situations in the dataset, and discuss computational
performance (Sections \ref{sec:discussion} and \ref{sec:terrain-mapping}).

\subsection{Experimental Platforms}
\label{sec:platforms}
The platforms used for our experiments are the ANYbotics ANYmal B300
\cite{Hutter2016} and C100 quadrupeds (\Figure \ref{fig:platforms-overview}).
Both robots have 4 identical legs giving a total of 12 active Degrees-of-Freedom
(DoF) and are equipped with an IMU and a Velodyne VLP-16 lidar. Each leg has
joint encoders, torque sensors. In some of the experiments, the robots were
modified from the stock version to compete in the DARPA SubT Challenge as
detailed in \cite{Tranzatto2021cerberus}. This gives a total of three different
sensor configurations (see \Figure \ref{fig:platforms-overview} and Table
\ref{tab:specs}). Note that, even though the experiments were performed using
the default leg configuration of ANYmal with inward pointing knees, VILENS is
agnostic to the number and configuration of legs used.

\begin{table}
\centering
\resizebox{\columnwidth}{!}{%
\begin{tabular}{llrl}
\toprule
\textbf{Sensor} & \textbf{Model} & \textbf{\si{\hertz}} &
\textbf{Specs} \\
\midrule
\midrule
\multicolumn{4}{l}{\textbf{Sensors common to ANYmal B300 and C100}}\\
 \midrule
Encoder & ANYdrive & 400 & \textit{Res:} \SI{<0.025}{\degree} \\
\midrule
Torque & ANYdrive & 400 & \textit{Res:} \SI{<0.1}{\newton\meter} \\
\midrule
Lidar & Velodyne VLP-16 & 10 & \textit{Res:} \SI{16 x 1824}{px} \\
\midrule
\midrule
\multicolumn{4}{l}{\textbf{ANYmal B300 (experiments: SMR, FSC)}}\\
 \midrule
\multirow{2}{*}{IMU} & Xsens & \multirow{2}{*}{400} &
\textit{Init Bias:} %
\SI{0.2}{\degree\per\second} $\vert$  \SI{5}{\milli\gram}  \\
 & MTi-100 &  & \textit{Bias Stab:} %
\SI{10}{\degree\per\hour} $\vert$  \SI{15}{\milli\gram}  \\
\midrule
Gray Stereo & RealSense &  \multirow{2}{*}{30} &  \textit{Res:} \SI{848 x
480}{px}  \\
Camera & D435i & & \textit{FoV (Diag.):} \SI{100.6}{\degree} \\
\midrule
 \midrule
 \multicolumn{4}{l}{\textbf{ANYmal B300 (experiment: SUB)}}\\
 \midrule
\multirow{2}{*}{IMU} & Xsens & \multirow{2}{*}{400} &
\textit{Init Bias:} %
\SI{0.2}{\degree\per\second} $\vert$  \SI{5}{\milli\gram}  \\
 & MTi-100 &  & \textit{Bias Stab:} %
\SI{10}{\degree\per\hour} $\vert$  \SI{15}{\milli\gram}  \\
\midrule
RGB Mono  & FLIR
 & \multirow{2}{*}{30} &  \textit{Res:} \SI{1440 x 1080}{px} \\
Camera & BFS-U3-16S2C-CS & & \textit{FoV (Diag.):} \SI{150}{\degree} \\
  \midrule
  \midrule
\multicolumn{4}{l}{\textbf{ANYmal C100 (experiments: LSM, SMM)}}\\
 \midrule
\multirow{2}{*}{IMU} & Epson & \multirow{2}{*}{400} &
\textit{Init  Bias:}
\SI{0.1}{\degree\per\second} $\vert$  \SI{3}{\milli\gram}  \\
 & G365 &  & \textit{Bias Stab:}%
 \SI{1.2}{\degree\per\hour} $\vert$  \SI{15}{\milli\gram}  \\
 \midrule
Gray Stereo & Sevensense &  \multirow{2}{*}{30} &  \textit{Res:} \SI{720 x
540}{px}  \\
Camera & Alphasense & & \textit{FoV (Diag.):} \SI{165.4}{\degree} \\
\bottomrule
\\
\end{tabular}
}
\caption{Sensor Specifications of the Experimental Platforms}
\label{tab:specs}
\vspace{-2mm}
\end{table}

\subsection{Dataset}
\label{sec:datasets}
To evaluate our proposed algorithm, we have collected datasets in a variety of
test environments lasting a total of \SI{2}{\hour} and traveling
\SI{1.8}{\kilo\meter}. This includes data from the Urban Circuit of the DARPA
SubT Challenge. The dataset is composed of the following experiments:
\begin{itemize}
\item\textbf{SMR:} Swiss Military Rescue Facility, Wangen, CH. Trotting over
concrete \& gravel followed by loops on grass with different gaits (ANYmal B300,
\SI{106}{\meter}, \SI{13}{\minute});
\item \textbf{FSC:} Fire Service College, Moreton-in-the-Marsh, UK. Three loops
of an outdoor industrial environment with standing water, oil residue, gravel
and mud (ANYmal B300, \SI{240}{\meter}, \SI{34}{\minute});
\item \textbf{SUB:} DARPA SubT Urban Beta course, Satsop, WA, USA\footnote{This
is the same sequence used in \cite{Wisth2020lidar}, except here we additionally
process leg kinematics.}. Autonomous exploration of a dark underground inactive
nuclear powerplant (ANYmal B300, \SI{490}{\meter}, \SI{60}{\minute});
\item \textbf{LSM:} A decommissioned limestone mine, Wiltshire, UK. Teleoperated
exploration of an unlit mine with several loops (ANYmal C100, \SI{474}{\metre},
\SI{20}{\minute});
\item \textbf{SMM:} \Seemuhle mine, CH. Autonomous exploration with Cerberus
SubT exploration system (ANYmal C100, \SI{522}{\metre}, \SI{17}{\minute}).
\end{itemize}

\begin{figure}
\centering
\includegraphics[width=0.49\columnwidth]{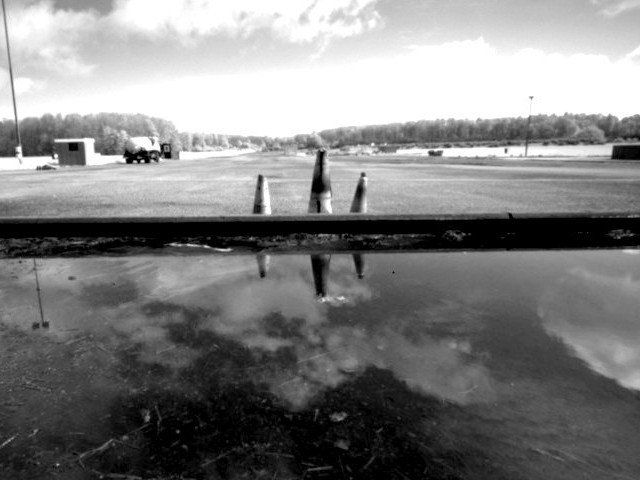}%
\hspace{0.01\columnwidth}%
\includegraphics[width=0.49\columnwidth]{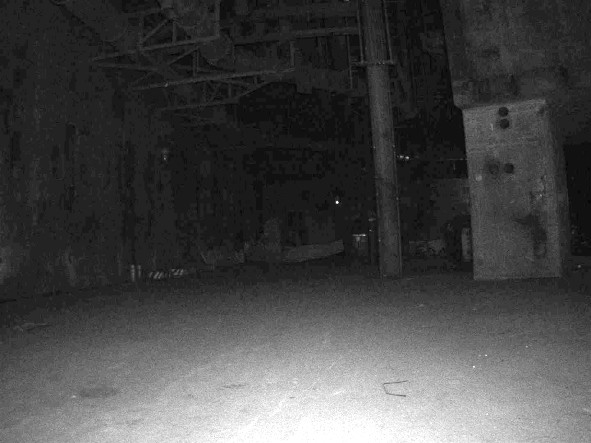} \\
\vspace{0.01\columnwidth}
\includegraphics[width=0.49\columnwidth]{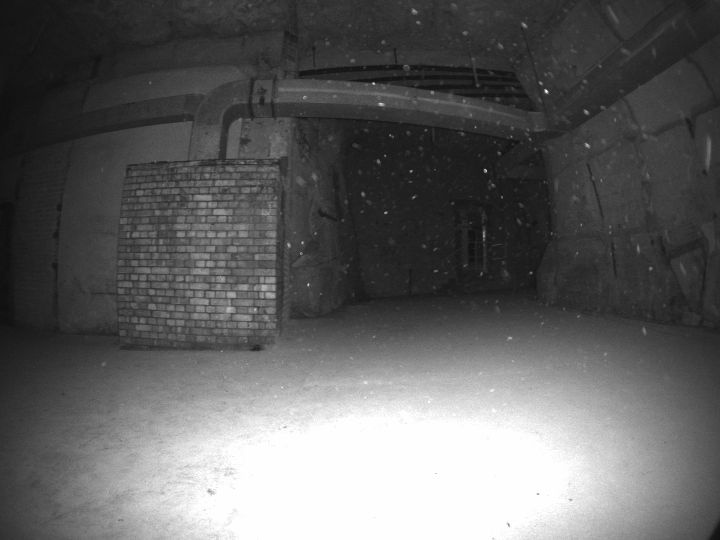}%
\hspace{0.01\columnwidth}%
\includegraphics[width=0.49\columnwidth]{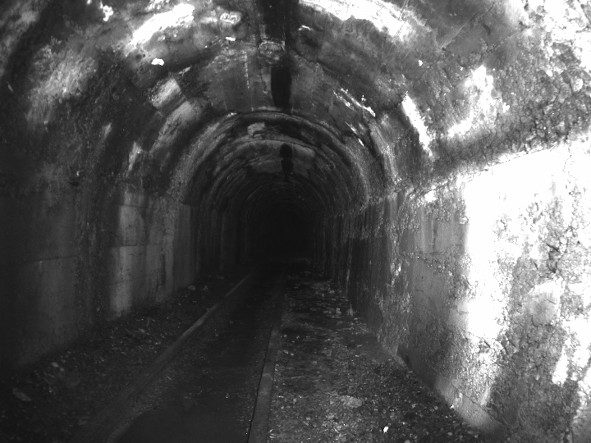}%
\caption{Onboard camera images showing challenging environments from the
experiments. \textit{Top-Left}: wet and oily concrete with reflections and lack
of features at the horizon (FSC); \textit{Top-Right:} complete darkness (image
manually enhanced in post processing). \textit{Bottom-Left:} low light,
overexposure and dust particles (LSM); \textit{Bottom-Right:} long straight
corridors and low light (SMM).}
\label{fig:scenarios}
\end{figure}

Images from the on-board cameras in each of the experiments are shown in \Figure
\ref{fig:scenarios} which illustrate various challenges including slippery
ground (affecting kinematics), reflections and darkness (affecting vision), and
long corridors (affecting lidar point cloud registration). Different copies of
the robots were used in each experiment.

In the SMR and FSC experiments, ground truth was generated by tracking the robot
using a Leica TS16 laser tracker and then estimating its orientation using a
optimization-based method (as described in \cite{Wisth2020icra}). In LSM, SUB,
and SMM experiments, an ICP-based method was to align the current lidar scan to
an accurate prior map made with survey-grade lidar scanners (as described in
\cite{Ramezani2020newer}).

\subsection{Multi-sensor Fusion Comparison}
\label{sec:multisensor-results}
To evaluate the performance of the system we compute the mean Relative Pose
Error (RPE) over distances of \SI{10}{\meter} traveled for different module
combinations of VILENS (at \SI{15}{\hertz}). We use a letter to indicate the
sensor modalities used as follows: lidar features (\textbf{L}), visual features
(\textbf{V}), IMU preintegration (\textbf{I}), preintegrated leg kinematics
(\textbf{K}), ICP registration (\textbf{R}).

The set of combinations tested are as follows:
\begin{itemize}
\item \textbf{VILENS-LVI:} lidar and visual feature tracking with preintegrated
IMU but no leg kinematics. This is the same configuration as
\cite{Wisth2020lidar}.
\item \textbf{VILENS-LVIK:} lidar and visual feature tracking, with
preintegrated IMU and including leg kinematics.
\item \textbf{VILENS-IR:} only ICP registration and preintegrated IMU. As a fair
comparison, we use the IMU to undistort point clouds. In this configuration, the
output is limited by the frequency of ICP and thus can only run at
\SI{2}{\hertz}.
\item \textbf{VILENS:} lidar and visual feature tracking, preintegrated IMU, leg
kinematics, and ICP registration. This is the full algorithm proposed in this
paper. (Where no letters are appended to the word VILENS we mean this version.)
\end{itemize}
Note that the same settings have been used for all configurations, with just
different modules activated for each experiment.

For comparison we also tested CompSLAM \cite{Khattak2020} which is a loosely
coupled filter combining lidar-inertial odometry (LOAM
\cite{zhang2017loam}), visual-inertial odometry (ROVIO \cite{Bloesch2017rovio}),
and kinematic-inertial odometry (TSIF \cite{Bloesch2017tsif}). It uses
heuristics to switch between modalities. LOAM is its primary modality and has a
frequency of \SI{5}{\hertz}. CompSLAM was also tested and deployed on the
ANYmal robots of the winning team of the DARPA Subterranean Challenge
\cite{Tranzatto2021cerberus}.

\begin{figure}
\centering
\includegraphics{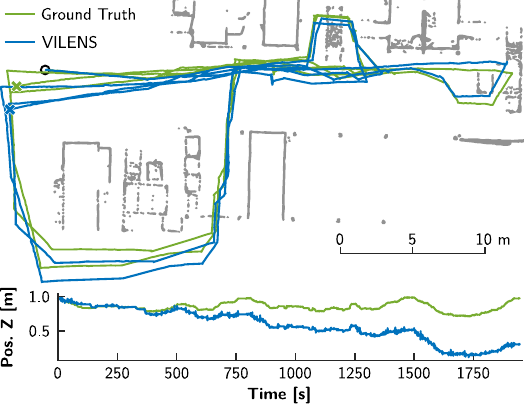}
\caption{Aerial view of the estimated and ground truth trajectories on the FSC
experiment (\SI{240}{\meter} traveled). The start of the trajectory is marked
with a circle. Ending location is marked with a cross. Note that since VILENS is
an odometry system, no loop closures have been performed.}
\label{fig:fsc-trajectory}
\end{figure}

\begin{table}
\centering
\resizebox{\columnwidth}{!}{
\begin{tabular}{l|ccccc}
\toprule
\multicolumn{6}{c}{\textbf{\SI{10}{\meter} Relative Pose Error (RPE)}} \\
 \toprule
 \multicolumn{6}{c}{\textbf{Translation	$\boldsymbol{\mu
(\sigma)}$[\si{\metre}]}} \\
\midrule
\textbf{Data} & \textbf{C-S}$^*$ \cite{Khattak2020} & \textbf{V-LVI}&
\textbf{V-LVIK}& \textbf{V-IR}$^\dagger$ & \textbf{VILENS}\\
\midrule
SMR & 0.28 (0.14) & 0.14 (0.12) & 0.15 (0.12) & 0.24 (0.15) & \textbf{0.12}
(0.11)\\
FSC & 0.16 (0.08) & 0.20 (0.09) & 0.24 (0.17) & 0.15 (0.09) &
\textbf{0.15} (0.07)\\
SUB & 0.20 (0.14) & 0.11 (0.07) & 0.11 (0.08) & 0.10 (0.08) & \textbf{0.05}
(0.03)\\
LSM & 0.27 (0.15) & 0.34 (0.30) & 0.29 (0.33) & 0.10 (0.08) & \textbf{0.04}
(0.04)\\
SMM & 0.36 (0.14) & 0.74 (0.99) & 0.74 (0.90) & 0.27 (0.22) & \textbf{0.12}
(0.08)\\
\midrule
Mean & 0.25 (0.13) & 0.31 (0.31) & 0.31 (0.32) & 0.17 (0.12) & \textbf{0.10}
(0.07) \\
\midrule
\multicolumn{6}{c}{\textbf{Rotation $\boldsymbol{\mu (\sigma)}$[\si{\degree}]}}
\\
\midrule
\textbf{Data} & \textbf{C-S}$^*$ \cite{Khattak2020} & \textbf{V-LVI}&
\textbf{V-LVIK}& \textbf{V-IR}$^\dagger$ & \textbf{VILENS} \\
\midrule
SMR & 3.14 (1.83) & \textbf{1.18} (0.97) & 1.38 (1.05) & 1.69 (0.93) &
1.30 (1.07)\\
FSC & 2.36 (1.07) & 1.30 (0.90) & 1.17 (0.94) & 2.03 (0.86) &
\textbf{1.14} (0.78) \\
SUB & 0.99 (0.62) & 0.75 (0.40) & 0.74 (0.43) & 1.88 (1.13) & \textbf{0.56}
(0.34)\\
LSM & 1.80 (0.94) & 2.32 (1.73) & 1.92 (1.51) & 1.44 (0.71) & \textbf{0.59}
(0.39)\\
SMM & 1.42 (0.73) & 4.73 (6.60) & 4.38 (5.26) & 2.73 (1.73) & \textbf{1.19}
(0.61)\\
\midrule
Mean & 1.94 (1.04) & 2.06 (2.12) & 1.92	(1.84) & 1.95 (1.07) & \textbf{0.96}
(0.64) \\
\bottomrule
\multicolumn{6}{c}{}  %
\end{tabular}
}
\caption{Experimental Results. ``C-S'' = CompSLAM, ``V-'' = VILENS-,
\linebreak $^*$output at \SI{5}{\hertz}, $^\dagger$output at \SI{2}{\hertz}}
\label{tab:multimodal-rpe}
\end{table}

\begin{figure}
\centering
\includegraphics{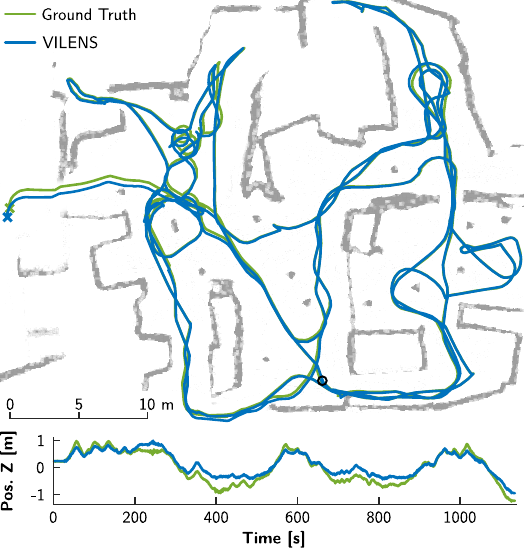}
\caption{Top-down view of the trajectory estimate by VILENS (blue) and the
ground truth (green) on the \SI{474}{\metre} LSM experiment. The start of the
trajectory is indicated by a black circle, while the end is indicated with a
cross. Note that since VILENS is an odometry system, no loop closures have been
performed.}
\label{fig:corsham-trajectory}
\end{figure}

\Figures \ref{fig:fsc-trajectory} and \ref{fig:corsham-trajectory} illustrate
the high level performance of VILENS. It can be seen that the estimated
trajectory closely matches the ground truth, showing the accuracy of this
approach. Note that VILENS is an odometry system --- no loop closures are
performed in this system. However VILENS can be integrated with an external SLAM
system such as \cite{Ramezani2020lidarslam}.

Quantitative results are summarized in Table \ref{tab:multimodal-rpe}. Across
the entire dataset, the best performing algorithm was the complete VILENS
system, with an overall mean RPE error of just \SI{0.96}{\percent} in
translation and \SI{0.0956}{\deg\per\meter} in rotation.

Compared to the ICP and IMU only (VILENS-IR) solution VILENS provides higher
frequency and higher accuracy estimation, as well as robustness against
degenerate scenarios for lidar, including long tunnels. The improvement in
performance is due to the incorporation of different sensor modalities at higher
frequency. This created a smoother state estimate which improved lidar motion
correction, provided a better prior for ICP, and allowed for inaccurate ICP
estimates to be rejected using robust cost functions.

Additionally, VILENS outperforms the loosely coupled approach, CompSLAM, by an
average of \SI{62}{\percent} in translation and \SI{51}{\percent} in rotation.

Accurate, high-frequency state estimation is also a requirement for terrain
mapping, as discussed in Section \ref{sec:terrain-mapping}.

\subsection{Twist Bias Ablation Study}
\label{sec:bias-resuts}

\begin{table}
\centering
\resizebox{\columnwidth}{!}{
\begin{tabular}{l|cc|cc}
\toprule
\multicolumn{5}{c}{\textbf{\SI{10}{\meter} Relative Pose Error (RPE)}} \\
 \toprule
 & \multicolumn{2}{c}{\textbf{Translation $\boldsymbol{\mu
(\sigma)}$[\si{\metre}]}} & \multicolumn{2}{c}{\textbf{Rotation
$\boldsymbol{\mu (\sigma)}$[\si{\degree}]}} \\
\midrule
\textbf{Data} & \textbf{VILENS-NO-BIAS} & \textbf{VILENS} &
\textbf{VILENS-NO-BIAS} & \textbf{VILENS}\\
\midrule
SMR & 0.12 (0.12) & \textbf{0.12} (0.11) & 1.30 (0.94) & \textbf{1.30} (1.07)\\
FSC & 0.21 (0.15) & \textbf{0.15} (0.07) & 1.24 (0.79) & \textbf{1.14} (0.78) \\
SUB & 0.06 (0.05) & \textbf{0.05} (0.03) & 0.64 (0.41) & \textbf{0.56} (0.34)\\
LSM & 0.05 (0.04) & \textbf{0.04} (0.04) & 0.64 (0.41) & \textbf{0.56} (0.34)\\
SMM & 0.13 (0.09) & \textbf{0.12} (0.08) & \textbf{1.12} (0.62) & 1.19 (0.61)\\
\midrule
Mean & 0.11 (0.09) & \textbf{0.10} (0.07) & 1.02 (0.65) & \textbf{0.96} (0.64)
\\
\bottomrule
\multicolumn{5}{c}{}  %
\end{tabular}
}
\caption{Ablation Study of Online Velocity Bias Estimation}
\label{tab:twist-bias-comparison-rpe}
\end{table}

In this section we quantitatively evaluate the effect of the velocity bias
estimation. As shown in Table \ref{tab:twist-bias-comparison-rpe}, even when
using all available sensor modalities (including cameras and lidar), the biased
signal from kinematics can still negatively effect the outcome. On average,
adding online bias estimation reduces the RPE on average by \SI{9.0}{\percent}
/ \SI{5.9}{\percent} in translation / rotation. However, the effect of bias
estimation is most obvious when exteroceptive sensors are degraded (see Section
\ref{sec:underconstrained-environments}).

\subsection{Internal Kinematics Comparison}
\label{sec:internal-kinematics}

\begin{figure*}
\centering
\includegraphics{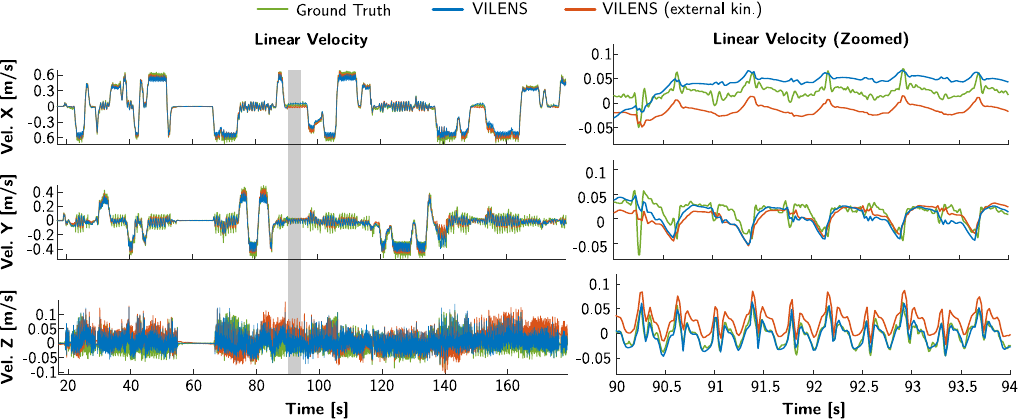}
\caption{\emph{Left:} Linear velocity comparison between VILENS fusing IMU and
leg kinematics using the external filter from
TSIF \cite{Bloesch2017tsif} (orange) and fusing the
kinematics inside the optimizer (blue) against ground truth (green).
\emph{Right:} zoomed in plot corresponding to the gray area of the complete
experiment. The better velocity tracking allows for a reduction in position
drift.}
\label{fig:int-vs-ext-kinematics-cov}
\end{figure*}

In this section we evaluate quantitatively the benefit of incorporating the
kinematics inside the factor graph, rather than relying on an external filter.
\Figure \ref{fig:int-vs-ext-kinematics-cov} shows a comparison between the
estimated velocity when using internal kinematics and receiving the
output from an
external kinematic-inertial filter \cite{Bloesch2017tsif}. The ground truth
was obtained by differentiation of the \SI{200}{\hertz} motion capture
data. Note that in this
configuration we are only using kinematic-inertial inputs (no exteroceptive
sensors) to highlight the difference between the two approaches. It can be seen
that using internal kinematics improves the velocity tracking, and in turn
reduces position drift by \SI{45}{\percent}.

\subsection{Integration with Local Planner and Perceptive Controller}
\label{sec:terrain-mapping}
This section will provide a qualitative demonstration of VILENS running
onboard the ANYmal robot. The limited on-board computation
is shared with other processes running on the
robot, including control, terrain mapping, and sensor drivers.

Constructing an accurate local terrain map around the robot is crucial for
perceptive locomotion and path planning. Locomotion controllers plan footstep
placements on these maps such as \cite{melon2021icra}.

The current approach for the ANYmal robot is to use the kinematic-inertial
estimator, TSIF \cite{Bloesch2017tsif}, to feed the local elevation mapping
system \cite{Fankhauser2018elevationmap}. TSIF can suffer from significant
drift, which creates ``phantom obstacles'' in the terrain map. The effect of
this is that the robot exhibits undesirable behavior such as poorly placed foot
landings or planning suboptimal paths. In the worst case, the robot may fall and
get irreversibly damaged or become stuck because the local path planner is
unable to find a feasible solution.

\begin{figure}
\centering
\includegraphics[width=0.9\columnwidth]{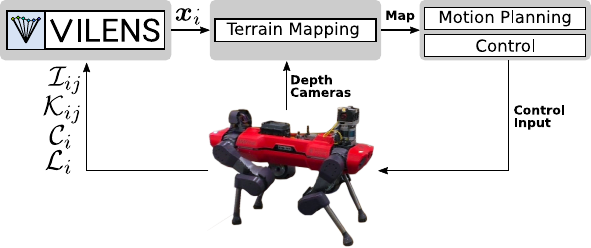}
\caption{\textit{Terrain Mapping Pipeline: } The VILENS state estimator produces
a high frequency (\SI{400}{\hertz}), low drift state estimate, $\State$, for the
terrain mapping module \cite{Fankhauser2018elevationmap}. This local terrain map
can then be used by other modules, such as perceptive motion planning and
control \cite{Gangapurwala2020}, or local path planning for obstacle avoidance
\cite{Dang2020gbplanner}.}
\label{fig:terrain-mapping-pipeline}
\end{figure}

\Figure \ref{fig:terrain-mapping-pipeline} shows the pipeline used for the
terrain mapping with VILENS. The inputs to the terrain mapping module
\cite{Fankhauser2018elevationmap} are the VILENS state estimate, and the point
clouds from several downward-facing depth cameras on the robot's body. These
cameras are not the same as the ones used for state estimation and are not
triggered at the same time. Therefore, accurate terrain mapping depends on a
smooth, accurate, and high-frequency state estimate to avoid interpolation
errors.

From our experience, terrain mapping frequencies of $\ge$\SI{15}{\hertz} are
required for dynamic locomotion over rough terrain. This means that
registration-based algorithms with a low frequency output (such as ICP or LOAM)
are not suitable for this purpose.

\begin{figure}
\centering
\includegraphics[width=0.493\columnwidth]{%
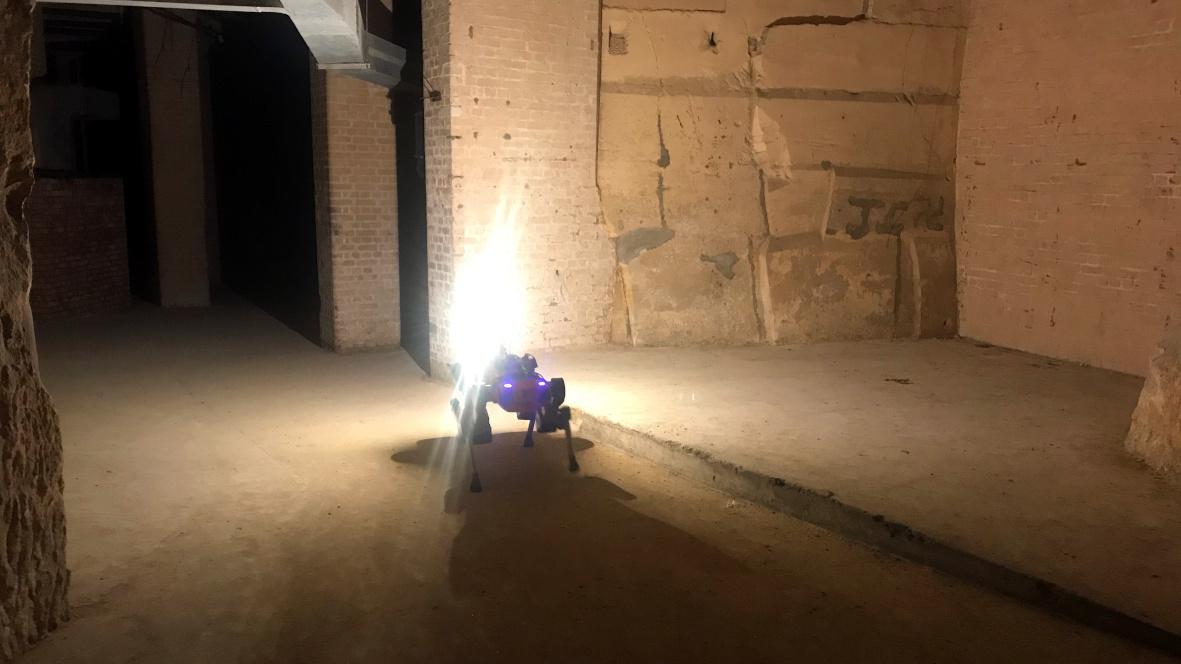}%
\includegraphics[width=0.493\columnwidth]{%
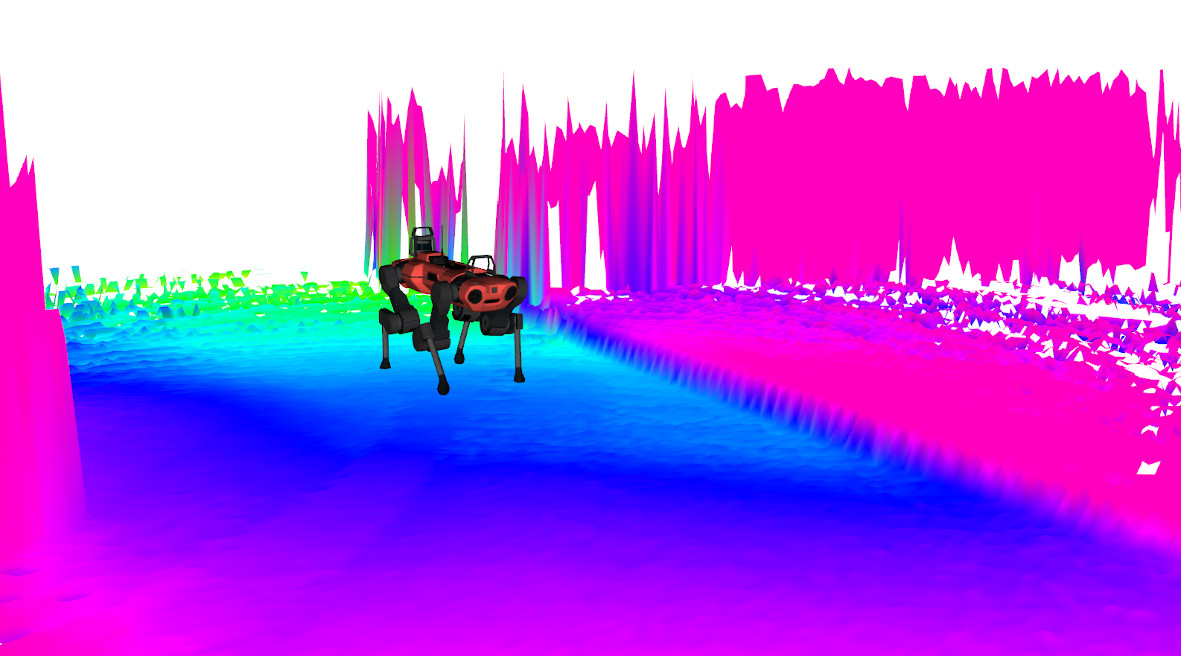}%
\caption{The accurate terrain map generated with the VILENS state estimate can
be fed into the local path planner \cite{Dang2020gbplanner} for autonomous
exploration. The ledge near the robot is clearly defined in the elevation map,
allowing the local planner to create routes on this obstacle.}
\label{fig:terrain-mapping}
\end{figure}

\subsubsection{Terrain Map Integration 1: Local Path Planning}
In the first experiment, we evaluate the quality of online local terrain mapping
for local path planning. During the LSM experiment the terrain map was
successfully used in the loop by local path planning algorithm, GBPlanner
\cite{Dang2020gbplanner}, for over \SI{30}{\minute} without failure. \Figure
\ref{fig:terrain-mapping}-\textit{Left} shows an example of the accuracy of the
terrain map compared to the ground truth.

\begin{figure}
\centering
\includegraphics[width=0.493\columnwidth]{%
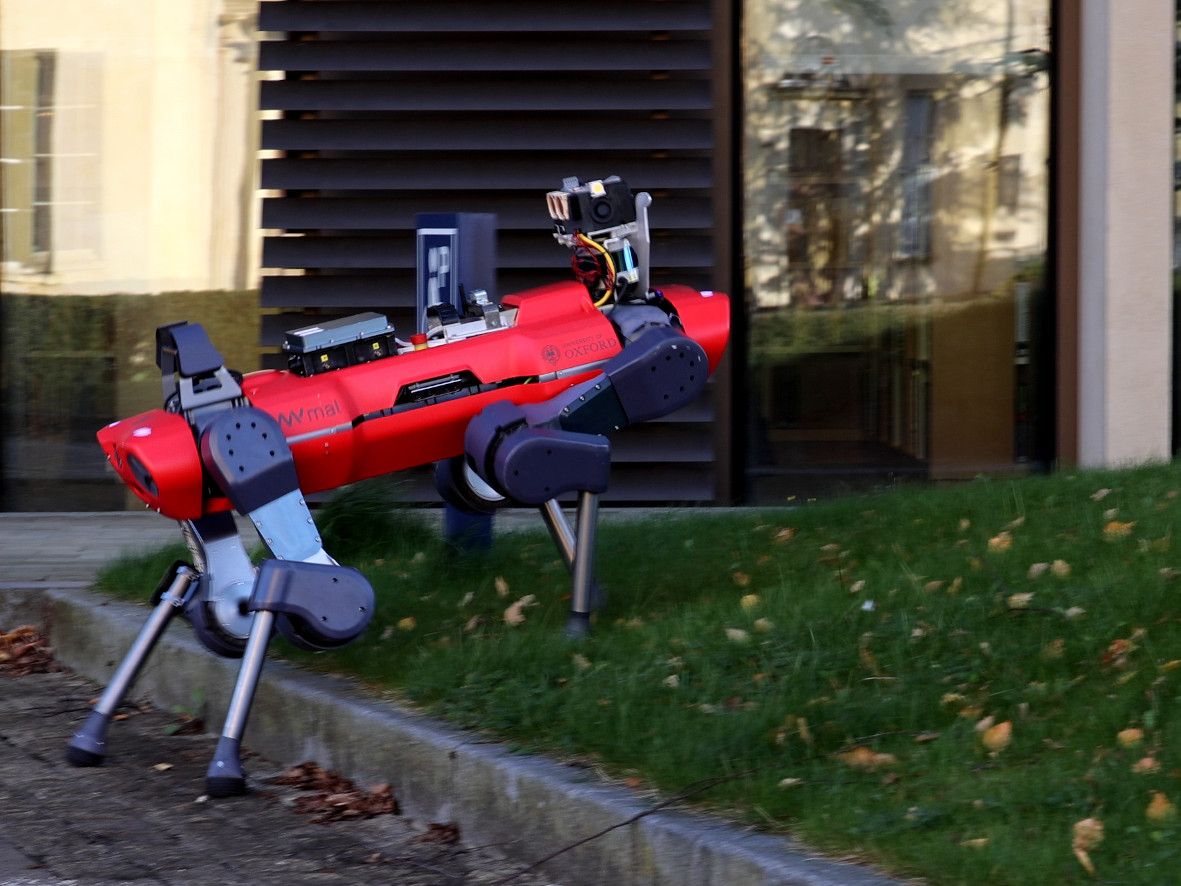}%
\hfill
\includegraphics[width=0.493\columnwidth]{
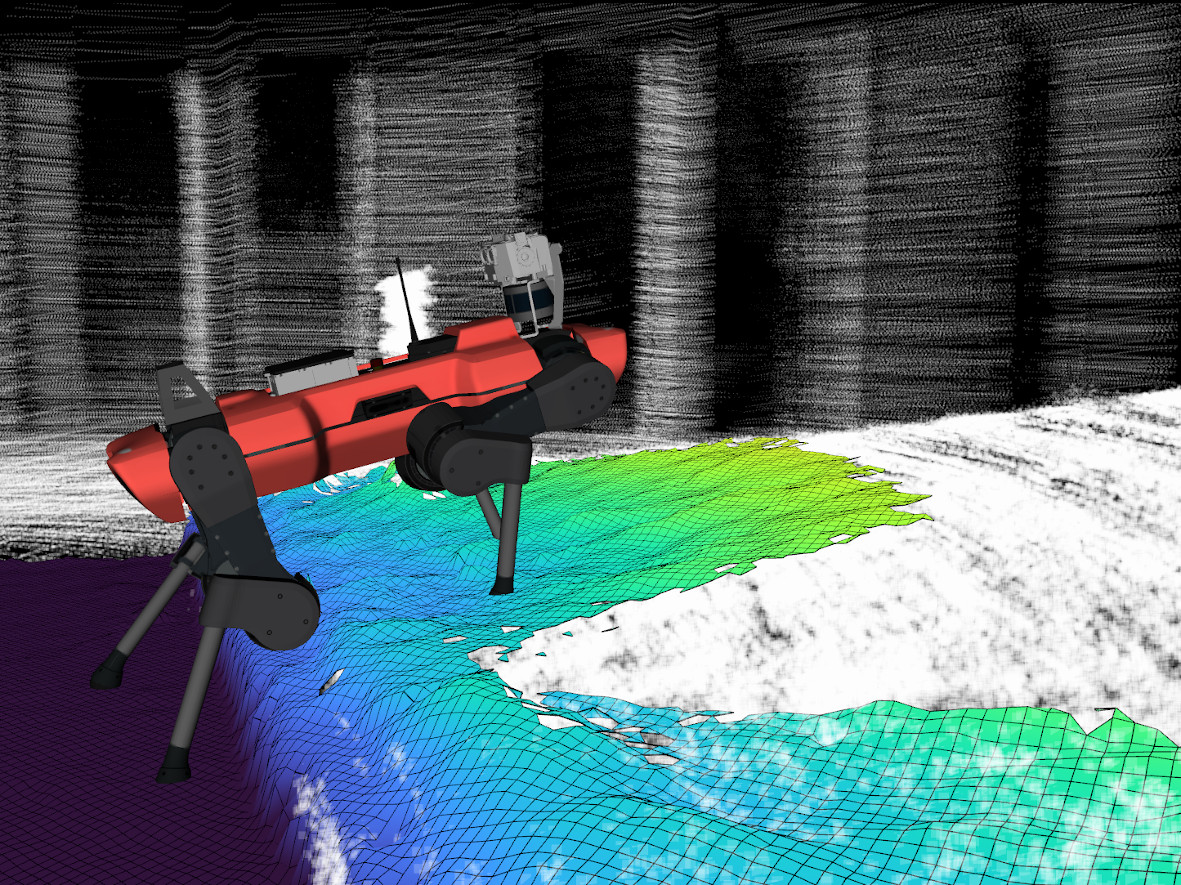}\\
\vspace{0.01\columnwidth}%
\includegraphics[width=0.493\columnwidth]{%
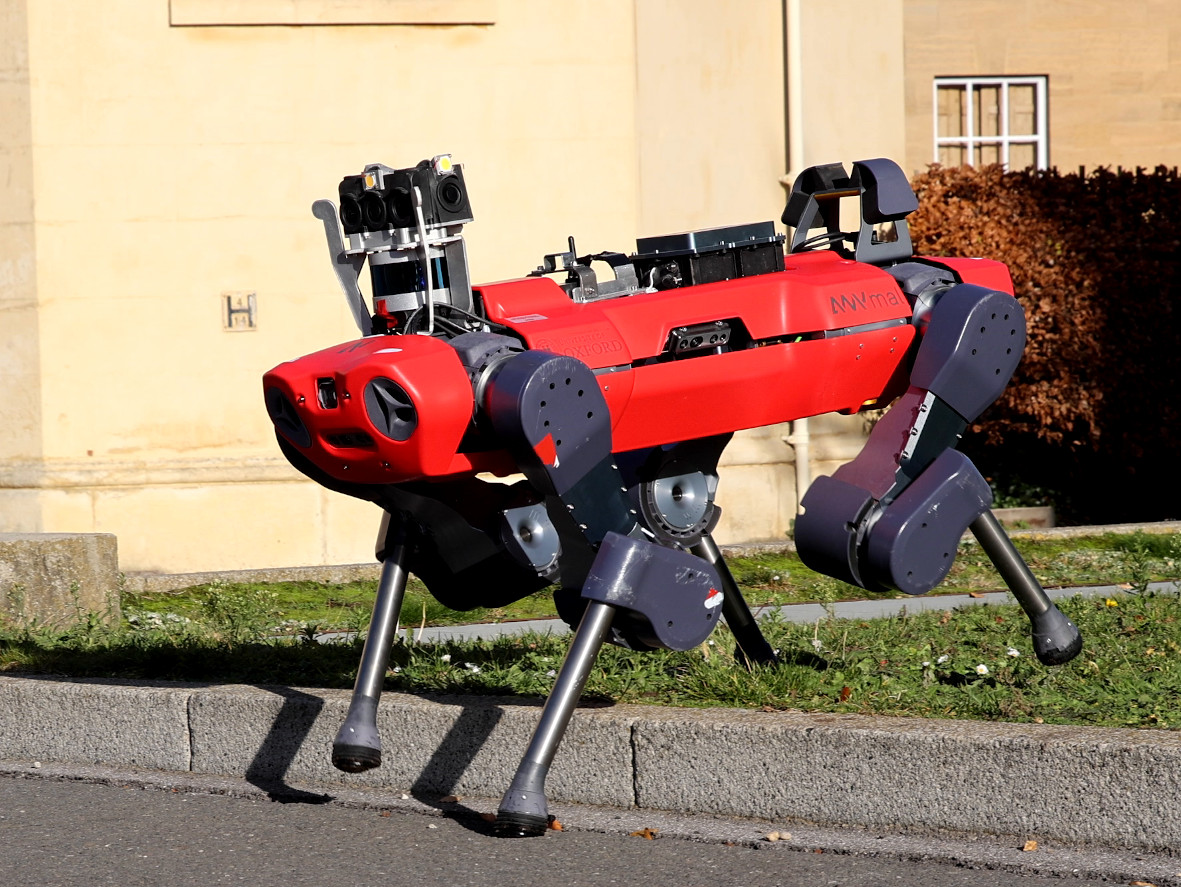} %
\hfill
\includegraphics[width=0.493\columnwidth]{%
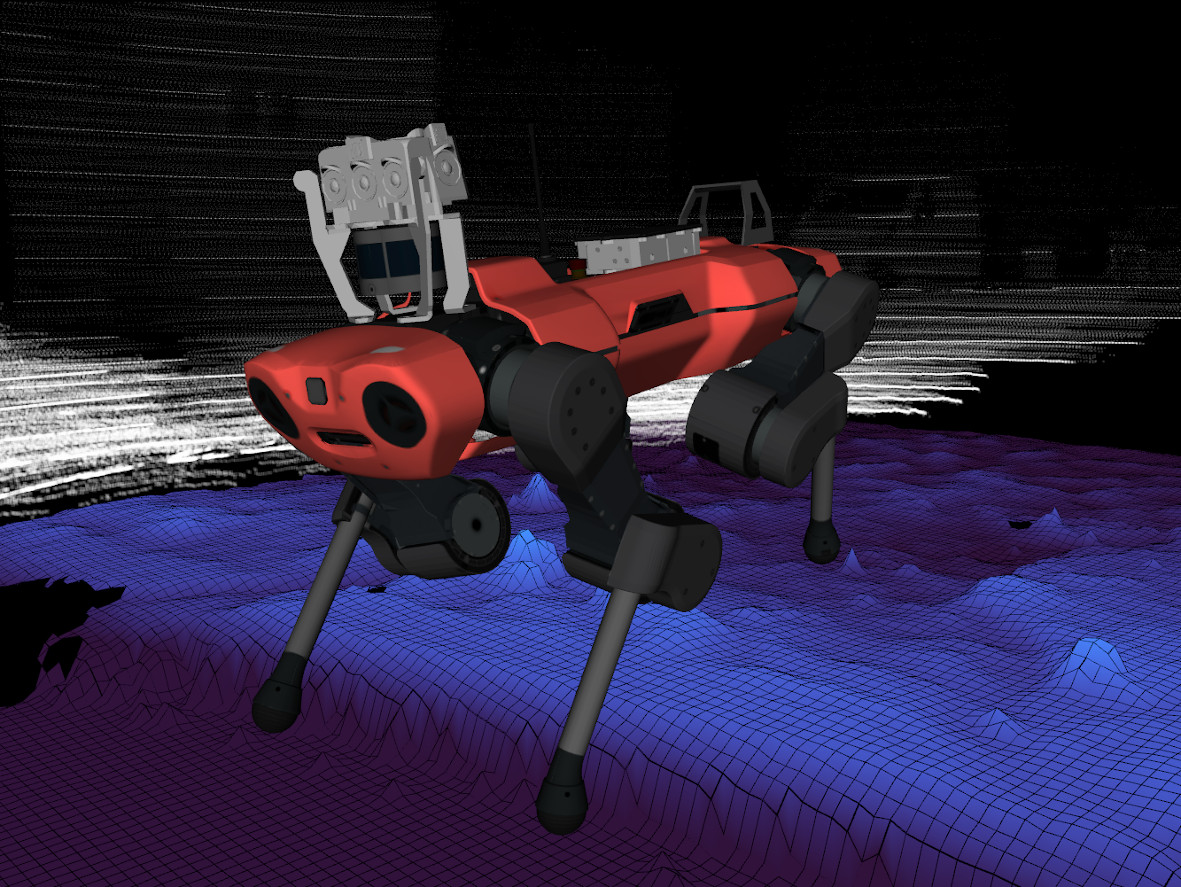} %
\caption{The ANYmal robot walking over curbs, slopes, and other rough terrain
using the RLOC perceptive controller \cite{Gangapurwala2020}. The accurate
terrain map produced by using the VILENS estimate enabled successful execution
of the trajectory.}
\label{fig:terrain-mapping-rloc}
\end{figure}

\subsubsection{Terrain Map Integration 2: Perceptive Control}
In the second experiment, we demonstrate terrain mapping for perceptive control
using RLOC \cite{Gangapurwala2020} during a \SI{5}{\minute} outdoor experiment
over grass, gravel, curbs, and slopes. \Figure \ref{fig:terrain-mapping} shows
the accuracy of the terrain map, which allowed the controller to plan and
execute precise steps on the terrain. By reducing the amount drift in the state
estimate, the quality of terrain map improves, allowing for more accurate motion
planning. Future work could involve tighter integration of local terrain mapping
and contact point locations estimated by VILENS.

\section{Discussion}
\label{sec:discussion}
\subsection{Evolution of Velocity Bias}
\label{sec:bias-evo}
\begin{figure}
 \centering
 \includegraphics{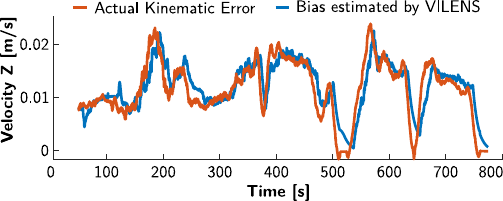}
 \caption{Comparison between the actual velocity error (from the kinematic
odometry in the $z$-axis inferred using a Leica tracker) and the bias estimated
by VILENS during the SMR experiment. Using exteroceptive sensing, VILENS is able
to accurately and stably track this effect by modeling the bias term $\bv$.}
\label{fig:bias-evo}
\end{figure}

In \Figure \ref{fig:bias-evo} we compare the actual and the estimated velocity
error from kinematics in the $z$-axis, which is where most of the drift
occurs for this particular sequence due to the presence of gravel. The orange
line indicates the error
between the true robot velocity and the one perceived by the leg odometry, while
the blue line shows the VILENS estimate of the same error, which is modeled as a
velocity bias term of the leg odometry factor, $\bv$. The sequence analyzed is
the same as the one shown in \Figure \ref{fig:position-drift-motivation}. The
high degree of correlation between the two signals demonstrates the
effectiveness of leg odometry velocity bias estimation.

\subsection{Performance in Underconstrained Environments}
\label{sec:underconstrained-environments}
In contrast to other recent loosely-coupled approaches to multisensor state
estimation \cite{Khattak2020, Palieri2020locus}, VILENS naturally handles
degenerate scenarios without requiring hard switches which are typically hand
engineered.

\begin{figure}
\centering
\includegraphics{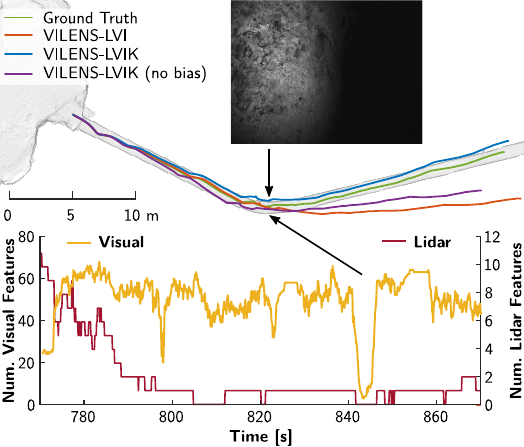}
\caption{\textit{Top:} Comparison between the trajectory estimates for different
VILENS configurations and ground truth (in green) for the tunnel sequence of the
SMM experiment. The lack of visual features causes a large rotation error when
not using leg odometry (orange line). Adding leg odometry improves the estimate
but still accumulates drift (purple line). With the online velocity bias
estimation, the drift is reduced even further (blue line). \textit{Bottom:}
number of lidar (in red) and visual (in yellow) features tracked over time. Note
that the lidar features plot has been slightly smoothed over a \SI{1}{\second}
window for clarity.}
\label{fig:seemuhle-degenerate}
\end{figure}

An example is shown in \Figure \ref{fig:seemuhle-degenerate}, where the ANYmal
robot walks very close to a wall in a long, straight tunnel. At time
\SI{843}{\second}, the robot turns towards the straight part of the tunnel. Due
to the degenerate geometry, the number of lidar features gradually drops as the
robot goes deeper into the tunnel (maroon line, bottom plot). While the lidar
feature tracking can still provide partial constraints to the system, ICP
is unstable and close to divergence, so it was not included into the analysis.
At the same time, the scene changes abruptly causing a period of underexposure
(picture at the top), with a dramatic drop in the number of visual features
(yellow line, bottom plot). Because of this, the VILENS-LVI configuration (\ie
VILENS without leg odometry and ICP, as in \cite{Wisth2020lidar}) accumulates a
large angular error (orange line). Instead, the addition of leg odometry factor
improves the estimate (VILENS-LVIK, blue). To show the benefit of velocity bias
estimation, we  also show the same VILENS-LVIK configuration but without online
bias estimation. Because the terrain is loose, the drift is accumulated faster
(purple line).

\subsection{Leg Odometry in Visually Degraded Scenarios}
\begin{figure}
\centering
\includegraphics{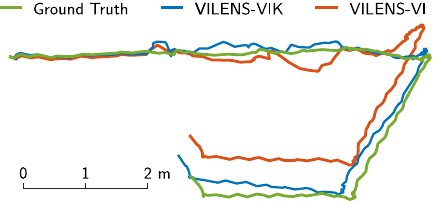}
\caption{Top-down comparison of VILENS-VIK and VILENS-VI trajectories aligned
with ground truth while crossing the puddle shown in \Figure \ref{fig:scenarios}
Top-Left.}
\label{fig:example_poor_vo}
\end{figure}

\Figure \ref{fig:example_poor_vo} shows a situation from FSC experiment where
the robot crosses a large puddle. VILENS-VI  tracks the features on the
standing water, causing drift on the $xy$-plane. Instead, VILENS-VIK maintains a
better pose estimate by relying on leg odometry with the bias estimation.

\subsection{Timing Analysis}
An important consideration in multisensor fusion algorithm is computation
time, which generally increases with the number of input sensors.

Table \ref{tab:timing} shows a summary of the computation time for the
components of VILENS. The timing tests were performed on a laptop equipped with 
an
Intel E-2186M processor (6 cores/12 threads @ \SI{2.9}{\giga\hertz} base
frequency) and 16 GB of RAM. A key benefit of using this type of factor-graph
based approach is that the lightweight visual and point cloud features allow for
accurate, low latency state estimation at relatively low computational expense.
This also allows more expensive modalities, such as ICP, to run at lower
frequency. This saves computation and preserves the benefits of low-drift ICP
state estimation.

\begin{figure}
\centering
\includegraphics{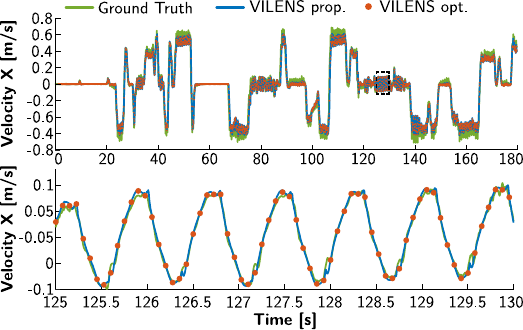}
\caption{\emph{Top:} Comparison between the optimized (\SI{10}{\hertz}, orange
dots) and IMU
forward propagated (\SI{400}{\hertz}, blue line) velocity estimates against the
ground truth from motion capture (green line). \emph{Bottom:} zoomed in view
corresponding to the dashed box at time interval \SIrange{125}{130}{\second}.
The close tracking of the ground
truth shows that IMU forward propagation provides an accurate, high-frequency,
and low-latency prediction of the state between optimizations.}
\label{fig:vilens-prop-vs-optimized}
\end{figure}

Table \ref{tab:output-timing} highlights the three different types of outputs
from the VILENS system, with differing levels of latency and accuracy for
different purposes. \Figure \ref{fig:vilens-prop-vs-optimized} shows how the
high-frequency forward propagated estimate closely matches Vicon ground truth.
This is important in some applications to capture the high-frequency behavior of
the robot between optimizations.

\begin{table}
\centering
\begin{tabular}{lcc}
\toprule
\textbf{Module} & \textbf{Timing $\mu(\sigma)$ [\SI{}{\milli\second}]} &
\textbf{Freq. [\SI{}{\hertz}]}\\
\midrule
IMU & 0.05 (0.12) & 400 \\
Leg kinematics & 0.07 (0.30) & 400 \\
Visual features with lidar depth & 9.48 (7.69) & 10 \\
Lidar point cloud features & 19.72 (6.24) & 10 \\
Lidar ICP & 149.75 (59.23) & 2 \\
\midrule
Optimization & 8.65 (3.25) & 10 \\
\bottomrule
\\
\end{tabular}
\caption{Analysis of timing for different parts of VILENS}
\label{tab:timing}
\end{table}

\begin{table}
\centering
\begin{tabular}{lcc}
\toprule
\textbf{Output type} & \textbf{Mean latency [\SI{}{\milli\second}]} &
\textbf{Freq. [\SI{}{\hertz}]} \\
\midrule
IMU forward-propagated & 2.3 & 400 \\
Factor graph optimized & 95.2 & 10 \\
ICP optimized & 395.2 & 2 \\
\bottomrule\\
\end{tabular}
\caption{Frequency and latency of outputs. Latency is compared to the lowest
latency input signal (IMU)}
\label{tab:output-timing}
\end{table}

\section{Conclusion}
\label{sec:conclusions}
This paper presented VILENS, a multisensor fusion algorithm which can
seamlessly fuse inertial, legged, lidar, and visual sensing within the same
factor graph. The tight fusion of all these sensor modalities allows the
algorithm to overcome the challenging operational conditions of autonomous
navigation in underground environments as well as large open areas. In these
conditions, individual sensor modalities such as visual-inertial or
lidar-inertial odometry would fail.

A particular contribution of the paper was the integration of the leg kinematic
measurements as a dedicated residual for the factor graph, rather than relying
on an external filter. This enables tighter integration and better noise
modeling. The leg odometry drift, typically occurring due to leg compression at
the impact, was modeled as a bias term of the velocity measurement and was
proven to be effective in challenging episodes where both the lidar and camera
sensors are deprived. In these situations, our system falls back to fusing IMU
and leg odometry with no hard switching required.

We demonstrated the robustness of our method in field experiments with several
quadruped platforms in challenging scenarios including slippery and deformable
terrain, water reflections, complete darkness, and degenerate geometries, such
as long corridors lasting longer than two hours.

The high frequency as well as the smoothness of the output was further
demonstrated by running the algorithm online and onboard the robot to enable
local elevation mapping, which was used by a perceptive controller
\cite{Gangapurwala2020} and a local path planner \cite{Dang2020gbplanner} to
cross an obstacle course and to autonomously explore a mine, respectively.

\appendix[Preintegrated Leg Odometry Factor]
\label{sec:appendix}
In this appendix we provide a derivation of the measurements and covariance for
the preintegrated leg odometry factor described in Section
\ref{sec:velocity-factor}. Without loss of generality, we will focus on cases
when only one leg is in contact with the ground. As explained in Section
\ref{sec:leg-odometry}, when there are multiple stance legs, the compound
measurement and noise can be computed independently from each stance leg and the
fused via a weighted average.

\subsection{Velocity Measurement and Noise}
\label{sec:derivation-leg-odo-velocity}
When a leg is in rigid, nonslipping contact with the ground, the robot's linear
velocity can be computed from the position and velocity of the feet:
\begin{equation}
\vel = - \Jacp(\jointpos)\jointvel - \rotvel \times \fkp(\jointpos)
\label{eq:legodo2}
\end{equation}
where $\fkp(\cdot)$ is the forward kinematics function and $\Jacp(\cdot)$ is its
Jacobian. To compute a velocity measurement for the factor graph, we need to
linearly separate the noise from the rest of \Equation \eqref{eq:legodo2}.
Remembering from \Equations \eqref{eq:joint-pos-noise} --
\eqref{eq:joint-vel-noise} that the joint states are corrupted by zero-mean
Gaussian noise, such that $\jointpos = \tjp - \etajp,\jointvel = \tjv - \etajv
$, we first start by separating the noise from the differential kinematics term:
\begin{align}
\Jacpp{\jp}\jv &=
\Jacpp{\tjp-\noise^{\alpha}}(\tilde{\jv}-\noise^{\dot{\alpha}}) \nonumber \\
 &= \Jacpp{\tjp-\noise^{\alpha}}\tilde{\jv} -
\Jacpp{\tjp-\noise^{\alpha}}\noise^{\dot{\alpha}} \nonumber \\
&\simeq \Jacpp{\tjp} \tjv - \partderivbig{\tjp}\left(\Jacpp{\tjp}\tjv
\right)\noise^{\alpha} - \Jacpp{\tjp}\noise^{\dot{\alpha}} \label{eq:jacjac}
\end{align}
where we applied the Taylor expansion of the product $\Jacpp{\jp}\jv$, ignoring
the second order terms. Since $\jp$ and $\jv$ are independent, $\jv$ can be
taken out of the derivative in \Equation \eqref{eq:jacjac}:
\begin{align}
\partderivbig{\tjp}\left(\Jacpp{\tjp}\tjv
\right)\noise^{\alpha} = \partderivbig{\tjp}\left(\Jacpp{\tjp}
\right)\tjv\noise^{\alpha} = \hessian{\tjp}
\tjv\noise^{\alpha}\label{eq:hessian}
\end{align}
where $\hessian{\cdot} \in \Real^{3 \times 3 \times 3}$ is the Hessian of the
forward kinematics function $\fkp(\jp)$. Note that, since the forward kinematics
is a vector function, the Hessian becomes a tensor of rank 3, so the product
with $\hessian{\tjp}\tjv = \hessian{\tjp}\, \bar{\times}_2 \,\tjv$ is a 2-mode
vector product \cite{KoBa09}. In our implementation, the
calculation of \Equation \eqref{eq:hessian} is performed numerically.

Substituting \Equations \eqref{eq:jacjac} -- \eqref{eq:hessian} into
\Equation \eqref{eq:legodo2} leads to:
\begin{multline}
\vel = - \Jacpp{\tjp} \tjv -\hessian{\tjp} \tjv\noise^{\alpha} -
\Jacpp{\tjp}\noise^{\dot{\alpha}} \\
- \rotvel \times \fkp(\tilde{\jointpos}-\noise^{\alpha})
\label{eq:step2}
\end{multline}
It has been demonstrated in \cite{Hartley2018b} that the noise from the last
term of \Equation \eqref{eq:step2} can be separated as follows:
\begin{equation}
\fkp(\tjp - \noise^{\alpha}) \approx \fkp(\tjp) - \Jacpp{\tjp}\noise^{\alpha}
\end{equation}
From the above relations, \Equation \eqref{eq:legodo2} becomes:
\begin{align}
\vel &= -(\Jacpp{\tjp} \tjv -\hessian{\tjp} \tjv\noise^{\alpha} -
\Jacpp{\tjp}\noise^{\dot{\alpha}}) \notag\\&\phantom{\hspace{0.55\columnwidth}}-
\rotvel^\wedge
\fkp(\tilde{\jointpos}-\noise^{\alpha}) \nonumber\\
&= -\Jacpp{\tjp} \tjv +\hessian{\tjp} \tjv\noise^{\alpha} +
\Jacpp{\tjp}\noise^{\dot{\alpha}} \notag\\&\phantom{\hspace{0.45\columnwidth}}-
\rotvel^\wedge
(\fkp(\tilde{\jointpos}) -\Jacpp{\tjp}\noise^{\alpha})\nonumber\\
&= -\Jacpp{\tjp} \tjv +\hessian{\tjp} \tjv\noise^{\alpha} +
\Jacpp{\tjp}\noise^{\dot{\alpha}} \notag\\&\phantom{\hspace{0.42\columnwidth}}-
\rotvel^\wedge
\fkp(\tilde{\jointpos}) + \rotvel^\wedge\Jacpp{\tjp}\noise^{\alpha}\nonumber\\
&= \underbrace{%
-\Jacpp{\tjp} \tjv- \rotvel^\wedge
\fkp(\tilde{\jointpos})}_{\tilde{\vel}}-
\notag\\&\phantom{\hspace{0.15\columnwidth}}\underbrace{\left(-\left(\hessian{
\tjp }
\tjv
+\rotvel^\wedge\Jacpp{\tjp}\right)\noise^{\alpha}-
\Jacpp{\tjp}\noise^{\dot{\alpha}}\right)}_{\etav} \nonumber%
 \\
 &\triangleq \tilde{\vel} - \etav
\label{eq:legodo-noise-out}
\end{align}
\Equation \eqref{eq:legodo-noise-out} still depends on the rotational velocity,
which is measured by the IMU and is affected by measurement noise and bias. If
we replace the angular velocity with the measured values, we have:
\begin{multline}
\vel = -\Jacpp{\tjp} \tjv- (\tw - \bw - \etaw)^\wedge
\fkp(\tilde{\jointpos}) + \\
\left(\hessian{\tjp} \tjv +(\tw - \bw -
\etaw)^\wedge\Jacpp{\tjp}\right)\noise^{\alpha} + \Jacpp{\tjp}\etav
\end{multline}
By applying the inversion rule for the cross product and the distributive
property we have:
\begin{multline}
\vel = -\Jacpp{\tjp} \tjv + \fkp(\tilde{\jointpos})^\wedge\tw -
\fkp(\tilde{\jointpos})^\wedge\bw - \fkp(\tilde{\jointpos})^\wedge\etaw
 + \\
\left(\hessian{\tjp} \tjv -\Jacpp{\tjp}^\wedge\tw + \Jacpp{\tjp}^\wedge\bw +
\Jacpp{\tjp}^\wedge\etaw\right)\etajp +\\
\Jacpp{\tjp}\etajv
\end{multline}
After rearrangement and removal of the second order terms, the noise can be
again separated as follows:
\begin{align}
\vel &\simeq \underbrace{-\Jacpp{\tjp} \tjv - {\tw}^{\wedge}\fkp(\tjp) +
{\bw}^\wedge\fkp(\tjp)}_{\tv} - \nonumber\\
&\quad\underbracea{\Big(-\left(%
\hessian{\tjp}\tjv+{\tw}^\wedge\Jacpp{\tjp}  -
{\bw}^\wedge\Jacpp{\tjp}\right)\etajp }  \nonumber\\
&\quad \underbraced{-\Jacpp{\tjp}\etajv  +
\fkp(\tilde{\jointpos})^\wedge\etaw\Big)}_{\etav} \nonumber\\
&\simeq \tv - \etav
\label{eq:noise-separation-final}
\end{align}
\subsection{Iterative Noise Propagation}
\label{sec:iter-noise-propagation}
By substitution of $\etav$ from \Equation \eqref{eq:noise-separation-final} into
\Equation \eqref{eq:preint-noise-model}, we can express the preintegrated noises
$\dVtheta, \dep$ in iterative form:
\begin{multline}
 \begin{bmatrix}
  \dVtheta_{i,k+1} \\
  \rhopert_{i,k+1}
 \end{bmatrix} =
  \underbrace{\begin{bmatrix}
  \dtTheta^\transpose_{k,k+1}  & \Zero \\
- \dtTheta_{i,k}(\tv_k - \bias^v_i)^\wedge \dt  & \Identity
  \end{bmatrix}}_{\Amat}
 \begin{bmatrix}
  \dVtheta_{i,k} \\
  \rhopert_{i,k}
 \end{bmatrix} + \\
  \underbrace{\begin{bmatrix}
    \JacR^k\dt  & \Zero  & \Zero \\
   -\fkp(\tjp_k)^\wedge  &
\boldsymbol{\chi}_k & \Jacp(\tjp_k)
  \end{bmatrix}}_{\Bmat}
\begin{bmatrix}
 \etaw \\
 \etajp\\
 \etajv
\end{bmatrix}
\label{eq:iterative-noise-expanded}
\end{multline}
where:
\begin{equation}
\boldsymbol{\chi}_k = \hessian{\tjp_k} \tjv_k +
{\twk}^\wedge \fkp(\tjp_k)-({\bw})^\wedge \Jacp(\tjp_k)
\end{equation}
\Equation \eqref{eq:iterative-noise-expanded} can be expressed more compactly
as:
\begin{equation}
 \noise^\Delta_{i,k+1} = \Amat \noise^\Delta_{i,k} + \Bmat \noise_k
 \label{eq:noise-compact}
\end{equation}
where $\noise_k = [ \etaw \;\etajp\; \etajv]^\transpose$.

From the linear expression \eqref{eq:noise-compact} and given the covariance
$\Cov_{\noise} \in \Real^{9\times9}$ of the raw gyro and joint states noises
$\noise_k$, the covariance for the factor can be computed iteratively:
\begin{equation}
 \Cov_{i,k+1} = \Amat \Cov_{i,k} \Amat^\transpose + \Bmat \Cov_{\noise}
\Bmat^\transpose
\end{equation}
starting from the initial condition $\Cov_{i,i} = \Zero$.

\subsection{Observability of the Linear Velocity Bias}
\label{sec:observability-analysis}

In \Equation \eqref{eq:vel-measurement} we have introduced the velocity bias
term $\bv$ to relax the non-slip relation from \Equation
\eqref{eq:no-slip}.
Intuitively, $\bv$
 represents the velocity of the contact point. If we rewrite \Equation
\eqref{eq:vel-measurement} and ignore Gaussian noise terms, the kinematic
velocity bias $\bv$ can be expressed as:
\begin{equation}
\bv = \tilde{\vel}_{kin} - \vel
\label{eq:convergence}
\end{equation}
where $\tilde{\vel}_{kin}$ is the estimated velocity from kinematics and $\vel$
is the true velocity.

\etalcite{Bloesch}{Bloesch2013} proved that, if the contact point is
stationary (\ie $\bv
= 0$), absolute position and
yaw are the only unobservable states (excluding degenerate cases). Therefore,
kinematic-inertial
measurements alone can only estimate $\tilde{\vel}_{kin}$.
The same unobservable states have been proved by
\etalcite{Hesch}{Hesch2014} for a typical visual-inertial system. Therefore, the
robot's linear velocity $\vel$ can be observed from IMU and camera measurements
only.

Since VILENS combines kinematic, inertial, and camera measurements, we speculate
that the linear velocity bias $\bv$ should be observable, as confirmed
empirically in our experiments. An analytic proof of observability and
convergence is left to future work.

\section*{Acknowledgment}
This research was part funded by the EU H2020 Projects THING and MEMMO the 
UKRI-funded ORCA Robotics Hub (EP/R026173/1). Fallon was supported by a Royal
Society University Research Fellowship and Wisth by a Google DeepMind
studentship. This research has been conducted as part of the ANYmal research
community.  Special thanks to the CERBERUS DARPA SubT Team for providing the
data from the challenge runs and the hardware support.

\ifCLASSOPTIONcaptionsoff
  \newpage
\fi

\addtolength{\textheight}{-0cm}  %

\balance

\bibliographystyle{./IEEEtran}
\bibliography{./IEEEabrv,./library}
\begin{IEEEbiography}
[{\includegraphics[width=1in,height=1.25in,clip,keepaspectratio]
	{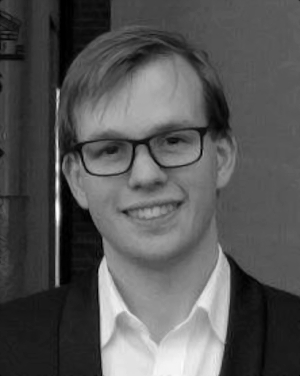}}]
{David Wisth} (Student Member, IEEE) received the B.S. degree in electrical
engineering in 2012 and the M.Eng. degree in mechatronics engineering in 2014
from the University of Melbourne, Australia. He is currently pursuing the
DPhil. degree in engineering science from the University of Oxford, UK.

From 2015 to 2017, he worked as a Project Engineer at Siemens, Australia. His
research interests include multisensor fusion, factor graphs, and legged
robots.

Mr Wisth's awards and honors include graduating as valedictorian from his
M.Eng. degree,  and Best Student  Paper Award (Finalist) at the IEEE
International Conference on Robotics and  Automation in 2021.
\end{IEEEbiography}
\begin{IEEEbiography}
[{\includegraphics[width=1in,height=1.25in,clip,keepaspectratio]
	{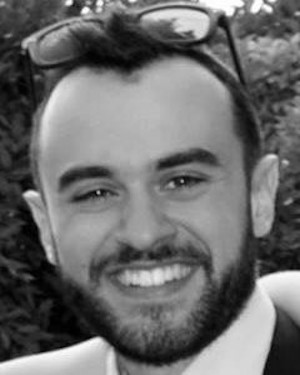}}]
{Marco Camurri} (Member, IEEE) received the B.Eng. and M.Eng. degrees in
Computer Engineering from the University of Modena and Reggio Emilia (Modena,
Italy) in 2009 and 2012, respectively. He received his PhD degree in Advanced
Robotics in 2017 from the Istituto Italiano di Tecnologia (IIT) in Genoa, Italy.

From 2017 to 2020 he was postdoc at IIT (one year) and University of Oxford (two
years).

Since 2021, he is Senior Research Associate at the Oxford Robotics Institute,
University of Oxford. His research interests include legged state estimation,
mobile robot perception, sensor fusion, mapping, and mobile autonomous
navigation.
\end{IEEEbiography}
\begin{IEEEbiography}
[{\includegraphics[width=1in,height=1.25in,clip,keepaspectratio]
	{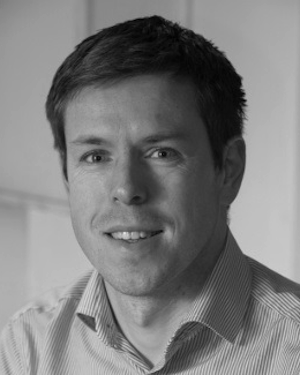}}]
{Maurice Fallon} (Member, IEEE) received his B.Eng (electrical engineering) from
University College Dublin, Ireland and his PhD (acoustic source tracking) from
University of Cambridge, UK.

From 2008--2012 he was a post-doc and research scientist in MIT Marine Robotics
Group working on SLAM. Later, he was the perception lead of MIT's team in the
DARPA Robotics Challenge. From 2017 he has been a Royal Society University
Research Fellow and Associate Professor at University of Oxford, UK. He leads
the Dynamic Robot Systems Group in Oxford Robotics Institute. His research
focuses on probabilistic methods for localization, mapping, dynamic motion
planning, and multisensor fusion.

\end{IEEEbiography}
\end{document}